\documentclass{article}

\usepackage{microtype}
\usepackage{graphicx}
\usepackage{subcaption}
\usepackage{booktabs} 
\usepackage{multirow} 
\usepackage{makecell} 
\usepackage{colortbl} 
\definecolor{mygray}{gray}{0.95} 

\usepackage{hyperref}

\usepackage[preprint]{icml2026}

\usepackage{amsmath}
\usepackage{amssymb}
\usepackage{mathtools}
\usepackage{amsthm}
\usepackage{enumerate}
\usepackage{paralist}
\usepackage{tikz}
\usetikzlibrary{shapes.geometric, arrows, positioning, calc}

\usepackage[capitalize,noabbrev]{cleveref}

\usepackage{multirow}
\usepackage[many]{tcolorbox}
\usepackage{makecell}
\tcbuselibrary{listings,breakable}
\usepackage{fontawesome5}
\tcbuselibrary{skins,breakable,raster}
\usepackage{marvosym}
\usepackage{xspace}
\usepackage{enumitem}
\usepackage{graphicx}
\theoremstyle{plain}

\theoremstyle{definition}

\theoremstyle{remark}

\usepackage[textsize=tiny]{todonotes}

\definecolor{simple_color}{HTML}{E38E80}    
\definecolor{stage2_color}{HTML}{E3A66F}    
\definecolor{explain_color}{HTML}{7AAAD6}   
\definecolor{reason_color}{HTML}{3E6A8A}    
\definecolor{question_color}{HTML}{3A3F4B} 
\definecolor{transfer_color}{HTML}{A1353A} 

\definecolor{correct}{RGB}{0, 150, 0}
\definecolor{wrong}{RGB}{150, 0, 0}
\definecolor{keytoken}{RGB}{180, 20, 20} 

\newtcolorbox{ExampleBox}[1][Example]{
    colback=gray!3,
    colframe=black!85,
    title=#1,
    fonttitle=\bfseries\normalsize,
    fontupper=\footnotesize,
    left=3pt,
    right=3pt,
    breakable
}
\newtcolorbox{SimpleBox}[1][Simple]{
    enhanced,
    breakable,
    colback=simple_color!8!white,
    colframe=simple_color,
    title=#1,
    fonttitle=\bfseries\small,
    fontupper=\scriptsize,
    left=4pt,
    right=4pt,
    top=3pt,
    bottom=3pt,
    arc=2mm,
    boxrule=0.5pt
}

\newtcolorbox{ExplainBox}[1][Explain]{
    enhanced,
    breakable,
    colback=explain_color!8!white,
    colframe=explain_color,
    title=#1,
    fonttitle=\bfseries\small,
    fontupper=\scriptsize,
    left=4pt,
    right=4pt,
    top=3pt,
    bottom=3pt,
    arc=2mm,
    boxrule=0.5pt
}

\newtcolorbox{ReasonBox}[1][Reasoning]{
    enhanced,
    breakable,
    colback=reason_color!8!white,
    colframe=reason_color,
    title=#1,
    fonttitle=\bfseries\small,
    fontupper=\scriptsize,
    left=4pt,
    right=4pt,
    top=3pt,
    bottom=3pt,
    arc=2mm,
    boxrule=0.5pt
}

\newtcolorbox{Stage2Box}[1][Stage2]{
    enhanced,
    breakable,
    colback=stage2_color!8!white,
    colframe=stage2_color,
    title=#1,
    fonttitle=\bfseries\small,
    fontupper=\scriptsize,
    left=4pt,
    right=4pt,
    top=3pt,
    bottom=3pt,
    arc=2mm,
    boxrule=0.5pt
}

\newtcolorbox{TransferBox}[1][Reason\ Transfer]{   
    enhanced,
    breakable,
    colback=transfer_color!8!white,   
    colframe=transfer_color,          
    title=#1,
    fonttitle=\bfseries\small,
    fontupper=\scriptsize,
    left=4pt,
    right=4pt,
    top=3pt,
    bottom=3pt,
    arc=2mm,
    boxrule=0.5pt
}

\definecolor{takeaway_color}{HTML}{D9A05B} 

\newtcolorbox{takeawaybox}[2]{
    enhanced,
    breakable,
    colback=takeaway_color!8!white,  
    colframe=takeaway_color,         
    title=#1,                        
    fonttitle=\bfseries\small,
    fontupper=\small,                
    left=4pt,
    right=4pt,
    top=3pt,
    bottom=3pt,
    arc=2mm,
    boxrule=0.5pt
}

\newcommand{\RightAns}{{\color{green!60!black} \faCheckCircle\ Right\ answer}}
\newcommand{\WrongAns}{{\color{red!70!black} \faTimesCircle\ Wrong\ answer}}

\makeatletter
\DeclareRobustCommand\onedot{\futurelet\@let@token\@onedot}
\def\@onedot{\ifx\@let@token.\else.\null\fi\xspace}
\makeatother

\def\eg{\emph{e.g}\onedot}

\def\ie{\emph{i.e}\onedot}

\def\etc{\emph{etc}\onedot}

\def \v {\mathbf{v}}
\def \t {\mathbf{t}}
\def \r {\mathbf{r}}
\def \p {\mathbf{p}}
\def \R {\mathbb{R}}

\newtheorem{prop}{Proposition}
\newtheorem{cor}{Corollary}

\icmltitlerunning{Rethinking Model Efficiency: Multi-Agent Inference with Large Models}

\begin{document}

\twocolumn[
\icmltitle{Rethinking Model Efficiency: Multi-Agent Inference with Large Models}

\begin{center}
\large 
Sixun Dong$^{1}$ \quad 
Juhua Hu$^2$ \quad 
Steven Li$^3$ \quad 
Wei Wen$^3$ \quad 
Qi Qian$^3$ $^\text{\Letter}$ \\
\vspace{2pt}
\normalsize
$^1$Independent Researcher, $^2$University of Washington, $^3$Meta Reality Labs\\
\vspace{1pt}
\texttt{\small sixundong.ai@gmail.com, juhuah@uw.edu, qiqian@meta.com}
\end{center}

\vskip 0.3in
]

\icmlcorrespondingauthor{Qi Qian}{qianq.mail@gmail.com}

\printAffiliationsAndNotice{}

\begin{abstract}
Most vision-language models (VLMs) apply a large language model (LLM) as the decoder, where the response tokens are generated sequentially through autoregression. Therefore, the number of output tokens can be the bottleneck of the end-to-end latency. 
However, different models may require vastly different numbers of output tokens to achieve comparable performance.
In this work, we conduct a comprehensive analysis of the latency across different components of VLMs on simulated data. The experiment shows that a large model with fewer output tokens can be more efficient than a small model with a long output sequence. The empirical study on diverse real-world benchmarks confirms the observation that a large model can achieve better or comparable performance as a small model with significantly fewer output tokens. 
To leverage the efficiency of large models, we propose a multi-agent inference framework that keeps large models with short responses but transfers the key reasoning tokens from the small model when necessary. The comparison on benchmark tasks demonstrates that by reusing the reasoning tokens from small models, it can help approach the performance of a large model with its own reasoning, which confirms the effectiveness of our proposal.

\end{abstract}   

\section{Introduction}
\label{sec:intro}

With the success of large language models (LLMs), vision-language models (VLMs) have been proposed to understand visual input with LLMs as the decoder~\cite{llava,qwen,internvl}. By inheriting the structure of LLMs, the output text token for VLMs is also generated in an autoregressive way, which requires a forward pass of the LLM for each token, and thus is expensive for inference.

To reduce the cost of inference, many small VLMs were developed to reduce the total number of parameters for inference. For example, SmolVLM~\cite{smolvlm} focuses on delivering models with a limited number of parameters (\eg, 500M and 256M), while showing promising performance on benchmarks. Meanwhile, leading VLM families also have small versions for efficiency. For example, InternVL3.5 series~\cite{internvl} provides 1/2/4B models, and Qwen3-VL family~\cite{qwen} also contains 2/4B models. Compared with their large-scale counterparts, those small ones can balance performance and the total number of parameters well.

\begin{figure}[tbp]
    \centering
    \captionsetup[subfigure]{skip=2pt}
    \begin{subfigure}{0.49\linewidth}
        \centering
        \includegraphics[width=\linewidth]{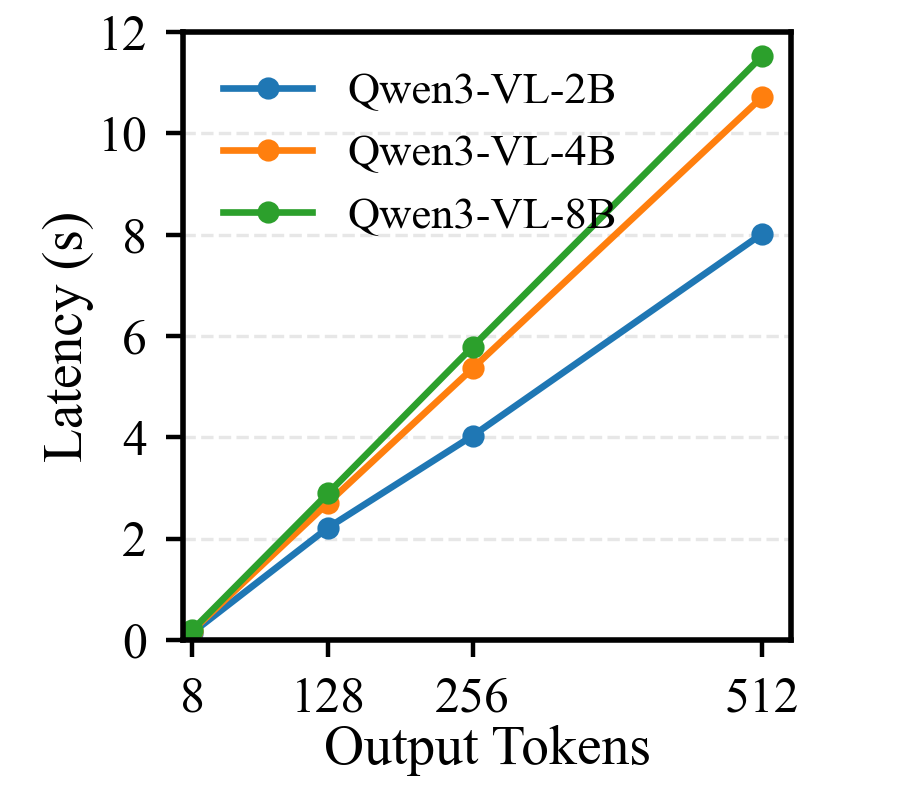}
        \caption{Latency vs. Output Tokens}
        \label{fig:running_time:left}
    \end{subfigure} 
    \hfill
    \begin{subfigure}{0.49\linewidth}
        \centering
        \includegraphics[width=\linewidth]{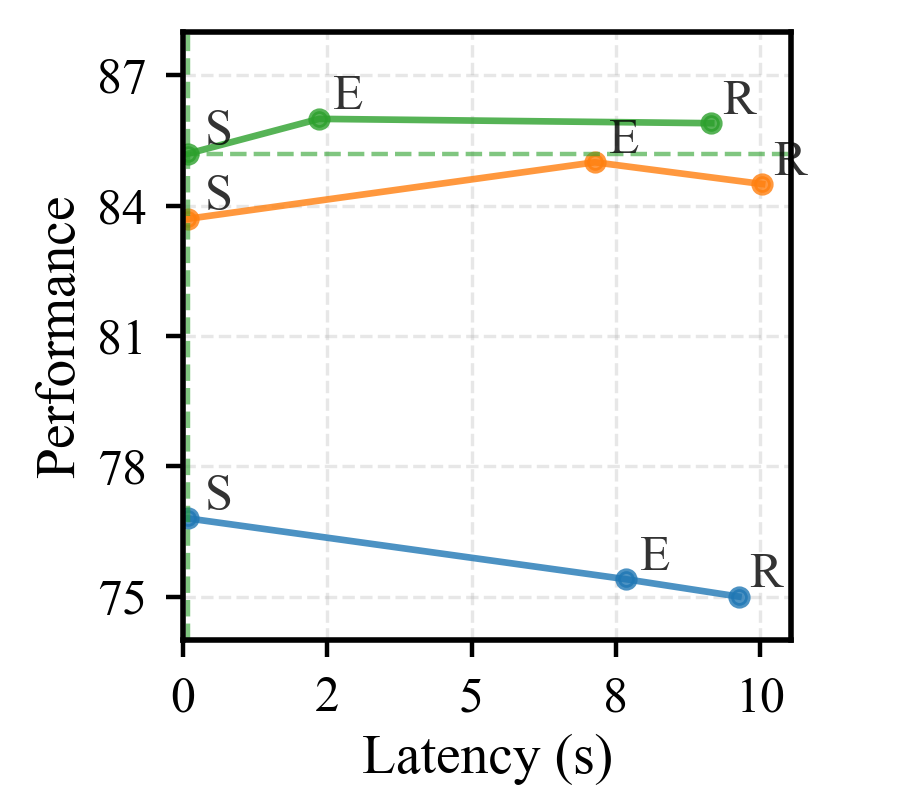}
         
        \caption{MMBench}
        \label{fig:running_time:right}
    \end{subfigure}
    
    \vspace{10pt}
    
     \begin{subfigure}{0.49\linewidth}
        \centering
        \includegraphics[width=\linewidth]{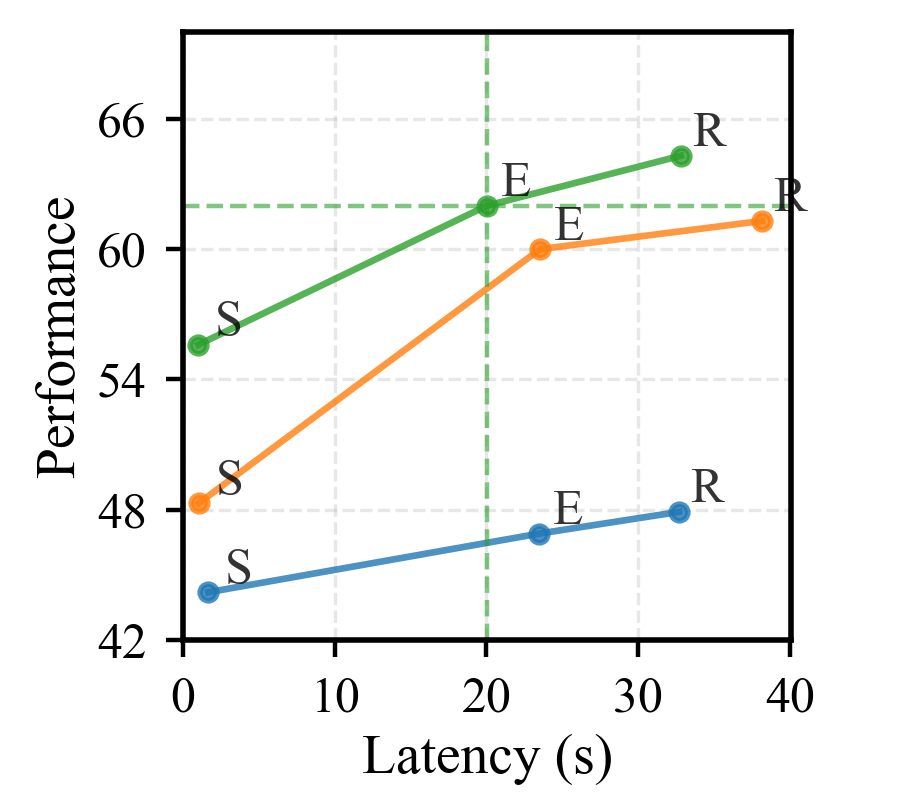}
         
        \caption{MMMU}
        \label{fig:running_time:right}
    \end{subfigure}
    \hfill
        \begin{subfigure}{0.49\linewidth}
        \centering
        \includegraphics[width=\linewidth]{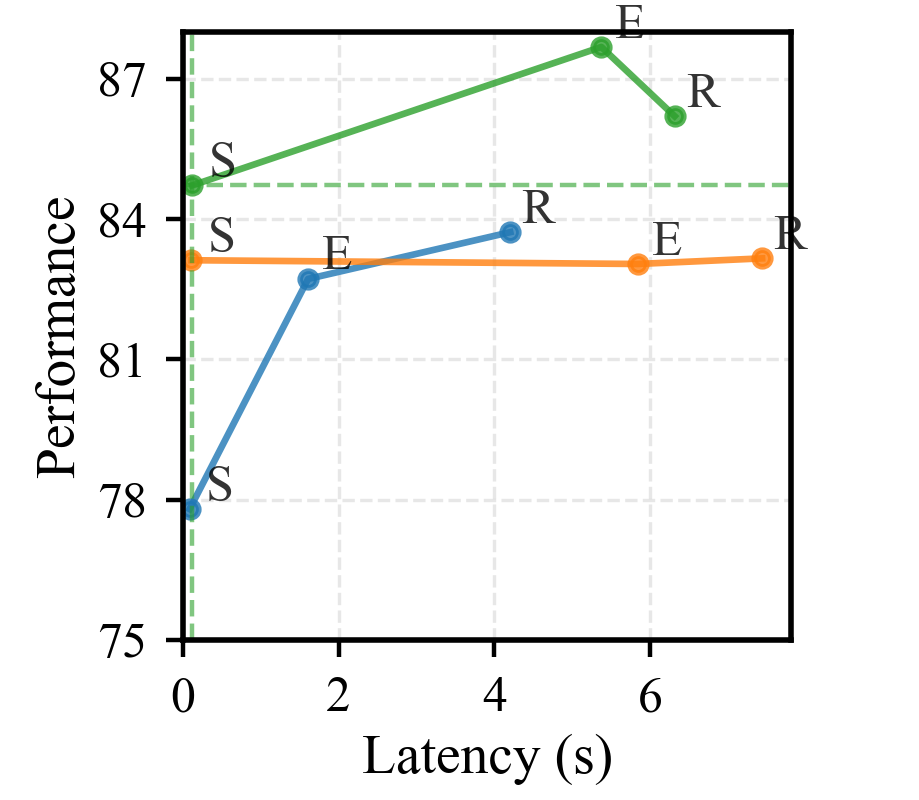}
        \caption{ChartQA}
        \label{fig:running_time:right}
    \end{subfigure}
    \caption{\textbf{Efficiency emerges with scale.} 
(a) Latency grows almost linearly on the number of output tokens, and larger models have the higher per-token cost.
(b)-(d) However, smaller models (2B/4B) require way more tokens to achieve a comparable performance as larger models (8B).}
    \label{fig:running_perf}
\end{figure}

In addition to the size of the model, the number of input tokens is also a critical factor contributing to the inference latency. This is because the essential block of the LLM in a VLM is the self-attention layer, whose computational cost is quadratic to the total number of input tokens. Moreover, visual input (\eg, images) usually produces a massive number of vision tokens with a high redundancy ratio. Therefore, visual token pruning is an active research area, where it shows that more than 90\% of visual tokens can be removed without significantly affecting performance~\cite{visionzip,mmtok}.

Although many efforts have been devoted to optimizing the size of models and the number of input tokens, they help only accelerate the inference of a single pass for an individual token. Since a complete response can contain multiple tokens, the end-to-end inference efficiency also depends on the total number of output tokens, which has been less investigated~\cite{WilhelmWK25}, especially for VLMs. In contrast, a thinking step can be included in response to improve quality~\cite{cot}, but requires more thinking tokens and thus increases latency significantly.

In this work, we aim to consider the inference efficiency from the perspective of output tokens. To isolate the influence of training receipt, we compare models with different sizes from the same model family. Given the same input for different models, we will compare performance and latency with different output tokens. First, with the simulation as in \cref{fig:running_perf} (a), we find that an 8B model with 128 output tokens can be more efficient than a 2B model with 256 output tokens. Due to the property of autoregressive generation, it is challenging to control the exact number of output tokens in real-world tasks. To mitigate this, we use different prompts to obtain answers with different scales of output tokens, \eg, Simple (S), Explain (E), Reasoning (R). As shown in \cref{fig:running_perf} (b)-(d), we observe that a large model, \eg, 8B, consistently achieves a better performance with fewer output tokens than the small model --\textbf{challanging the conventional assumption that small models are more inference-efficient}. To leverage the efficiency of large models with short output sequences, we propose a multi-agent framework with mutual verification and reasoning transfer.
Our framework can enjoy the efficiency of large models with fewer tokens and that of small models with more reasoning tokens. 
The main contributions of this work can be summarized as follows.
\begin{itemize}[noitemsep, topsep=0pt]
\item We apply 2/4/8B models from Qwen3-VL~\cite{Qwen3VL_GitHub_2025} and 1/2/4/8/14B models from InternVL3.5~\cite{internvl} with different numbers of output tokens on diverse benchmark data sets for a comprehensive comparison about end-to-end inference efficiency.
\item We observe that the degeneration performance of small models can come from the challenge of instruction following for the format of evaluation. A 2-stage strategy can improve small models substantially.
\item Considering the decoding latency, we show that a large model with fewer output tokens can be more efficient than a small model with reasoning in terms of achieving a targeted performance.
\item According to these observations, we propose a multi-agent inference framework that can transfer the selected reasoning tokens from small models to large models after mutual verification, which helps balance the performance and number of output tokens.
\end{itemize}

\section{Related Work}
\label{sec:related}

In the literature, many efforts have been devoted to improving the inference efficiency of VLMs as follows.

\subsection{Model Size Optimization}
Recent research has devoted significant efforts to improving the inference efficiency of vision-language models (VLMs) through model compression and scaling. SmolVLM~\cite{smolvlm} introduces compact multimodal models with parameter numbers ranging from 256M to 500M, while maintaining competitive accuracy on various benchmarks. Similarly, large-scale VLM families such as InternVL3.5~\cite{internvl} and Qwen3-VL~\cite{qwen} offer smaller variants (1B–4B) that balance performance and computational cost via distillation. Despite these advances, most efforts focus on reducing the number of model parameters rather than optimizing output-side efficiency—an aspect that this work specifically investigates.

\subsection{Input Visual Token Pruning}
Another line of work aims to accelerate inference by reducing the number of input tokens, particularly in the visual encoder. Since the computational complexity of the transformer’s self-attention grows quadratically with the total token count, pruning redundant vision tokens can significantly reduce latency. Techniques such as patch merger~\cite{qwen}, and importance-based selection~\cite{mmtok,visionzip} demonstrate that about 90\% of visual tokens can be safely removed without significant performance degradation. These approaches are complementary to model compression, as they optimize the input representation side. However, even with aggressive pruning, the inference still scales linearly with the number of output tokens—a factor rarely examined before our study.

\subsection{Reasoning with Small Models}

Considering the per-token efficiency of small models, some work also tries to leverage reasoning tokens from small models to help large models~\cite{leviathan2023fast,scot,liu2025small}. For example, \cite{scot} fine-tunes a small draft model to generate multiple candidate reasoning trails for the large model. \cite{liu2025small} proposes a similar strategy but uses multiple small models to obtain candidate reasoning paths for large models. \cite{offloaded} also offloads thinking tokens to small models while the application is limited to reasoning models. More recently, speculative reasoning methods~\cite{specreason,speccot} accelerate chain-of-thought by drafting and verifying reasoning segments, while multi-model collaboration approaches coordinate models through token-level routing~\cite{r2r} or external thought injection~\cite{liu2025thought}. However, most of these methods operate at the tight token level and often require specific training. In contrast, our framework operates entirely at the response level, making it training-free and natively compatible with optimized serving engines like vLLM. Notably, these two paradigms are complementary: existing token-level speculative decoding techniques could be seamlessly integrated within each model call in our framework to further reduce per-call latency.
\section{Number of Output Tokens Matters}

\subsection{Preliminaries}
Popular VLMs, including InternVL~\cite{internvl} and QwenVL~\cite{qwen}, first use a vision encoder $V$ to convert visual input $I$ to vision tokens as
\begin{eqnarray}
[\v_1,\dots, \v_m] = V(I)
\end{eqnarray}
Meanwhile, tokens for the query $Q$, can be extracted by a text tokenizer $T$
\begin{eqnarray}
[\t_1, \dots, \t_n] = T(Q)
\end{eqnarray}
where $\forall i, j, \v_i, \t_j\in\R^d$.

Given the vision and query tokens as input, the first response token can be obtained from the LLM after the vision encoder as
\begin{eqnarray}
r_1 = LLM([\v_1,\dots, \v_m, \t_1, \dots, \t_n])
\end{eqnarray}
Then, the response will be generated in an autoregressive manner that predicts the next token according to all input tokens and previously generated tokens
\begin{eqnarray}
r_s = LLM([\v_1,\dots, \v_m, \t_1, \dots, \t_n,\r_1, \dots, \r_{s-1}])
\end{eqnarray}
The generation process will stop when a special token is obtained, \eg, \texttt{<eot>}.

According to this process, it is obvious that the end-to-end latency of token generation from VLMs depends on 
\vspace{-3mm}
\begin{enumerate}
\setlength{\itemsep}{0pt}
  \setlength{\parsep}{0pt}
  \setlength{\parskip}{0pt}
    \item The number of parameters in the vision encoder $V$
    \item The number of vision tokens from the input image: $m$
    \item The number of text query tokens: $n$
    \item The number of parameters in $LLM$
    \item The number of output tokens: $s$
\end{enumerate}
\vspace{-3mm}
Although the first four factors are key for a single decoding pass, the total number of output tokens indicates how many trials are needed for decoding and is crucial for the inference efficiency.

\subsection{VLM Inference Profile: Small vs. Large}
We start our analysis with a detailed profiling of 2/4/8B Qwen3-VL models. According to the demonstration above, the inference process can be divided into four parts: input preprocessing, vision encoder, prefilling, and decoding. 
After warm-up processes, we report the running time of each part that is averaged over 100 runs using simulated data on a single H100.

\noindent \textbf{Input Preprocessing} Since we compare models from the same Qwen family, the input preprocessing pipeline is almost identical across models, and the latency is mainly from the input image size. As shown in \cref{tab:qwen_vision_pre}, we can find that there is no significant difference between different model sizes, and we will omit this in the following discussion to simplify the latency analysis.

\begin{table}[htbp]
    \centering
    \caption{Qwen3-VL vision preprocessing latency (ms) with different image resolutions.}
    \label{tab:qwen_vision_pre}
    \resizebox{0.9\linewidth}{!}{
    \begin{tabular}{c|ccccc}
        \toprule
        Qwen3-VL  & 256$^2$ & 512$^2$ & 1024$^2$ & 2048$^2$ & 4096$^2$ \\
        \midrule
        2B & 0.84 & 1.49 & 7.21 & 64.83 & 335.19 \\
        4B & 0.91 & 1.69 & 6.99 & 76.63 & 343.78 \\
        8B & 0.92 & 1.58 & 6.46 & 75.59 & 341.61 \\
        \bottomrule
    \end{tabular}
    }
    
\end{table}

\noindent \textbf{Vision Encoder} The vision encoder is a transformer model that converts image patches to tokens. According to the Qwen design, the 2B and 4B models share the same architecture of the vision encoder (\ie, SigLIP2-Large (300M)), while the 8B model has a larger vision encoder (\ie, SigLIP2-SO-400M). We also notice the difference by our profiling in \cref{tab:qwen_vision_encoder}. Compared to the 2B model, the 8B model can be about $60\%$ slower with an extremely large input size, \eg, $4096\times 4096$. However, for other large resolutions, the gap is not that large. For example, with $1024\times 1024$ as the input size, the difference between 2B and 8B is about 10ms, which is negligible. This observation implies that large models are still efficient with appropriate input resolutions, while small models are more efficient given very large input resolutions. The exploration of the trade-off between model size and image resolution is left for further work.
 
\begin{table}[htbp]
    \centering
    \caption{Qwen3-VL vision encoder latency (ms) with different image resolutions.}
    \label{tab:qwen_vision_encoder}
    \resizebox{0.9\linewidth}{!}{
    \begin{tabular}{c|ccccc}
        \toprule
        Qwen3-VL & 256$^2$ & 512$^2$ & 1024$^2$ & 2048$^2$ & 4096$^2$ \\
        \midrule
        2B & 12.69 & 12.84 & 24.41 & 172.28 & 1995.25 \\
        4B & 12.36 & 12.56 & 24.91 & 180.10 & 2104.70 \\
        8B & 13.79 & 14.26 & 34.40 & 272.90 & 3343.02 \\
        \bottomrule
    \end{tabular}
    }
    
\end{table}

\begin{table}[htbp]
    \centering
    \caption{{Qwen3-VL} Time-to-First-Token latency (ms) with different image resolutions and query token lengths.(Profiled to 8192 tokens, 1536 shown for brevity.)}
    \label{tab:qwen_ttft}
    \resizebox{\linewidth}{!}{
    \begin{tabular}{c|ccccccccc}
        \toprule
        Image Res. & \multicolumn{3}{c}{$256^2$ (64 Tokens)} & \multicolumn{3}{c}{$512^2$ (256 Tokens)} & \multicolumn{3}{c}{$1024^2$ (1024 Tokens)} \\
        \cmidrule(lr){1-2}\cmidrule(lr){2-4} \cmidrule(lr){5-7} \cmidrule(lr){8-10}
        Query Tokens & 64 & 256 & 512 & 64 & 256 & 512  & 64 & 256 & 512  \\
        Total Tokens & 128 & 320 & 576 & 320 & 512 & 768 & 1088 & 1280 & 1536 \\
        \midrule
        2B & 27.52 & 27.79 & 27.92 & 28.22 & 28.00 & 28.89 & 45.96 & 46.97 & 49.80 \\
        4B & 32.21 & 32.15 & 34.66 & 32.79 & 32.86 & 39.02 & 65.87 & 69.33 & 77.86 \\
        8B & 33.76 & 36.25 & 47.93 & 36.74 & 43.91 & 53.45 & 94.27 & 102.65 & 115.31 \\
        \bottomrule
    \end{tabular}
    }
    
\end{table}
\begin{table*}[tbp]
    \centering
    \caption{Qwen3-VL decoding time with different image resolutions and query token lengths.}
    \label{tab:qwen_decoding}
    \resizebox{\linewidth}{!}{
    \begin{tabular}{c|cccc|cccc|cccc|cccc|c}
        \toprule
        Image Res. & \multicolumn{8}{c|}{$256^2$ (64 Image Tokens)} & \multicolumn{8}{c|}{$512^2$ (256 Image Tokens)} & \multirow{3}{*}{Avg Tok/s} \\
        \cmidrule(lr){1-2}\cmidrule(lr){2-9}\cmidrule(lr){10-17}
        Input Tok. & \multicolumn{4}{c}{128} & \multicolumn{4}{c|}{320} & \multicolumn{4}{c}{320}  & \multicolumn{4}{c|}{512} & \\
        \cmidrule(lr){2-5}\cmidrule(lr){6-9}\cmidrule(lr){10-13}\cmidrule(lr){14-17}
        Output Tok. & 8 & 128 & 256 & 512 & 8 & 128 & 256 & 512 & 8 & 128 & 256 & 512 & 8 & 128 & 256 & 512 &  \\
        
        \midrule
        Qwen3-VL-2B  & 138.73 & 2030.41 & 4051.51 & 8072.48 & 139.68 & 2032.67 & 4052.28 & 8070.89 & 139.62 & 2038.54 & 4056.94 & 8080.48 & 138.45 & 2021.54 & 4035.18 & 8067.07 & 61.80 \\
        Qwen3-VL-4B  & 176.63 & 2636.46 & 5275.21 & 10515.65 & 177.41 & 2642.43 & 5266.63 & 10486.47 & 177.36 & 2644.82 & 5267.35 & 10487.88 & 176.38 & 2634.05 & 5261.63 & 10503.87 & 47.77 \\
        Qwen3-VL-8B  & 178.50 & 2651.51 & 5282.48 & 10517.45 & 180.89 & 2656.08 & 5300.42 & 10540.37 & 181.69 & 2660.82 & 5294.05 & 10563.77 & 186.50 & 2648.22 & 5271.76 & 10690.40 & 47.26 \\
        \bottomrule
    \end{tabular}
    }
    
\end{table*}

\noindent \textbf{Prefilling} To show the prefilling latency, we report Time to First Token (TTFT), which accounts for the latency of a single forward pass and consists of four parts: input preprocessing, vision encoding, prefilling, and generating the first output token. The result is summarized in \cref{tab:qwen_ttft}. Although the 8B model is 4 times larger than the 2B model in terms of parameters, the latency of large models increases slightly. Even with the large input resolution (\ie, $1024\times 1024$) and $512$ query tokens, the TTFT of the 8B model is only about 2 times larger than the 2B model. With a small image resolution (\eg, $256\times 256$) and limited query tokens (\eg, 64), the 8B model only costs an additional 6ms to obtain the first tokens compared to the 2B model, which shows the efficiency of large models. This observation inspires us to explore the potential of large models for obtaining a complete response by decoding multiple tokens.

\noindent \textbf{Decoding} Now we conduct a comparison on different numbers of generated tokens to demonstrate the efficiency of different models in real scenarios. According to \cref{tab:qwen_decoding}, we can find that the number of output tokens dominates the overall latency. Although the 2B model can generate eight tokens within 140ms, it costs about 2000ms to obtain 128 tokens, which incurs more latency than any other factors discussed above. Compared with the 2B model, the 4B and 8B models are a bit slower in terms of averaged output tokens per second (Tok/s). However, the gap is mild and much smaller than that of the model size. Even with an 8B model, it can produce up to 47 tokens per second compared to 61 from a 2B model. This phenomenon confirms that the number of output tokens is essential for the latency of VLMs. With a different number of output tokens, a large model with 128 output tokens can be faster than a small model with 256 output tokens, as shown in \cref{fig:running_perf} (a). 

Moreover, as shown in \cref{fig:running_perf} (b), the performance of large models with fewer output tokens can also be better than that of small ones with more tokens. Given these insights from our comparison, we propose a multi-agent inference framework to explore the inference efficiency with small and large models sufficiently.

\begin{takeawaybox}{}{}  
\includegraphics[width=0.8em]{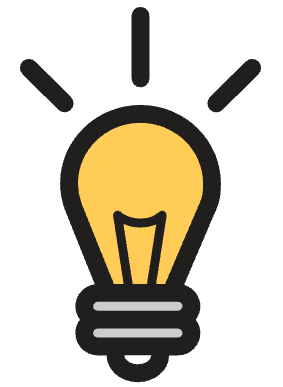}~\textbf{Takeaway 1:} 
Decoding cost dominates end-to-end VLM latency. Since the per-token cost gap across model scales is mild (61.8 vs.\ 47.3 tok/s for 2B vs.\ 8B) while the output token gap is $2\times$, a larger model with fewer tokens can be faster than a smaller model with verbose reasoning.
\end{takeawaybox}

However, large models with simple prompts are not always sufficient. As illustrated in Appendix ~\ref{sec:append_example} (Example 3), when both 8B-S and 2B-R fail on a challenging query, only 8B-R yields the correct answer. This motivates a key question: can the cheap reasoning tokens from the small model substitute for the expensive reasoning of the large model?

\section{Multi-Agent Inference: Mutual Verification and Reasoning Transfer} \label{sec:mai}

Let $VLM_l$ and $VLM_s$ denote a large VLM and a small VLM, respectively. Since both models show small latency with short output sequences, we start with mutual verification.

\subsection{Mutual Verification} 
First, we ask two models to answer the query directly without any reasoning, which can limit the number of output tokens.
\begin{eqnarray}
&&A_{short}^l = VLM_l(\text{Q+Prompt$_{simple}$})\\
&&A_{short}^s = VLM_s(\text{Q+Prompt$_{simple}$})
\end{eqnarray}
where $A_{short}^l$ and $A_{short}^s$ are short answers from the large and small models, respectively. Prompt$_{simple}$ is the prompt to let the model answer the query without thinking.

We extract key answers via standard parsing protocols (e.g., LMMs-Eval) and determine semantic equivalence using Exact Matching or ANLS (as per benchmark), ensuring the verification process incurs negligible computational overhead.

If $A_{short}^s$ is semantically identical to $A_{short}^l$, there is an agreement between large and small models, and the final result is obtained with a limited number of output tokens.

When two responses are different, the models may need some thinking tokens to figure out the right answer. To avoid the long sequence inference from large models, we rely on small models for reasoning. However, we empirically observe that small models may not follow the output format after too much thought. To mitigate the issue, we decompose the process into two stages: reasoning + answer as follows.
\begin{align}\label{eq:2stage}
&R_{think}^s = VLM_s(\text{Q+Prompt$_{think}$})\\
&A_{think}^s = VLM_s(\text{Q+$R_{think}^s$+Prompt$_{simple}$})
\end{align}
where Prompt$_{think}$ is the prompt that allows the model to think before answering, and $R_{think}^s$ is the corresponding output thinking tokens from a small model and can be a long sequence. After that, $R_{think}^s$ is attached as the contextual prompt for the small model to answer the question again in $A_{think}^s$. With an appropriate KV cache, the gap to a single inference with a long output sequence should be small, but the instruction following can be better as shown in our experiments. 

Now we have another answer from the small model, and we will verify it with the output from the large model again. If $A_{think}^s = A_{short}^l$, it will be returned as the final output. However, if those two answers are still different, we may need reasoning from large models.

\begin{figure*}
    \centering
    \includegraphics[width=1\linewidth]{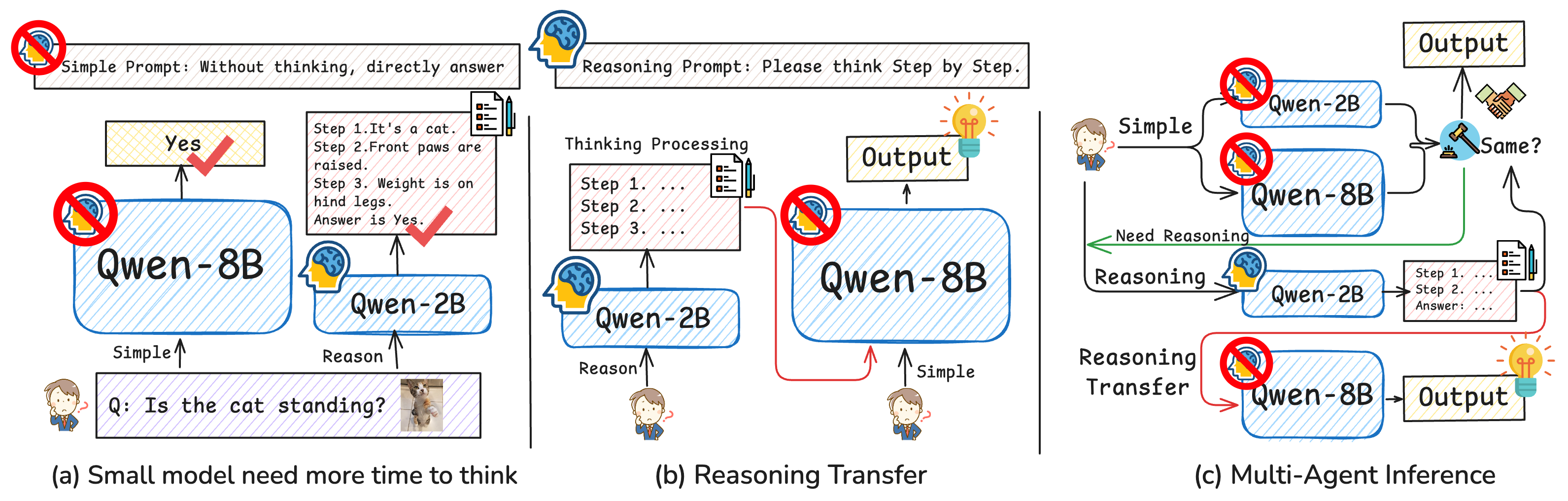}
    \caption{Illustration of the proposed multi-agent inference framework. (a) shows our empirical observation that a large model with a short response can achieve a similar performance as the small model with additional reasoning tokens. (b) demonstrates the proposed reasoning transfer strategy that can reuse the reasoning tokens output by the small model for the large model to improve its performance. (c) Our final proposal adopts mutual verification to further reduce the number of expensive model calls for efficient inference.}
    \label{fig:placeholder}
\end{figure*}

\subsection{Reasoning Transfer} 
Although the large model is efficient for short responses, it can be slow for reasoning by generating a lot of thinking tokens.  Since prefilling is substantially cheaper than decoding (\cref{tab:qwen_ttft} vs. \cref{tab:qwen_decoding}), we propose to transfer the reasoning process from the small model to the large model to avoid the costly sequential decoding entirely.

Concretely, we reuse the thinking tokens $R_{think}^s$ from the small model as the contextual prompt for the large model and let it answer the question referring to the analysis of the small model. 
\begin{align} \label{eq:rt}
&A_{think}^l = VLM_l(\text{Q+$R_{think}^s$+Prompt$_{simple}$})
\end{align}

Compared with thinking from the large model directly, it saves the cost of generating thinking tokens iteratively
\begin{align}
&R_{think}^l = VLM_l(\text{Q+Prompt$_{think}$})
\end{align}

However, $R_{think}^l$ can be different from $R_{think}^s$, and $R_{think}^s$ can contain noise for large models. Fortunately, thanks to self-attention, the attention pattern can be sparse and focus on a few useful tokens automatically.

Let $X_l\in\R^{n\times d}$ and $X_s\in\R^{n\times d}$ denote the token representations of $R_{think}^l$ and $R_{think}^s$ respectively. $d$ is the dimension of the features and $n$ is the number of tokens. Without losing generality, we assume that the number of tokens is the same for $R_{think}^l$ and $R_{think}^s$, which can be easily achieved by padding. Since the answer is generated by the prediction of the next token, we will investigate the last token from the prompt. Note that the target token itself is omitted from $X_l$ and $X_s$ for demonstration.

We find that the difference after attention can be bounded with sparse entities.

\begin{cor}\label{cor:1}
Let 
\begin{align}
&\p_l = \texttt{softmax}(\p X_l^{\top}/\lambda)X_l;\p'_l = \texttt{softmax}(\p X_l^{'\top}/\lambda)X'_l\nonumber\\
&\p_s = \texttt{softmax}(\p X_s^{\top}/\lambda)X_s;\p'_s = \texttt{softmax}(\p X_s^{'\top}/\lambda)X'_s\nonumber
\end{align}
$X'_l$ and $X'_s$ are subsets of $X_l$ and $X_s$ respectively, where the attention score $\sum_{j:j\in X'_l}  q_l^j\geq \delta, \sum_{j:j\in X'_s} q_s^j\geq \delta$. $\delta$ is a positive constant in $[0.8,1]$. By assuming the norm of tokens is bounded, \ie $\forall i\quad \|X_l^i\|_2\leq c, \|X_s^i\|_2\leq c$, we have
\begin{eqnarray}
\|\p_l - \p_s\|_2 \leq \|\p'_l - \p'_s\|_2 + 4(1-\delta)c
\end{eqnarray}
\end{cor}

The detailed proof can be found in \cref{sec:append_proof}. Corollary~\ref{cor:1} shows that the difference of representations after the self-attention layer mainly depends on the dominating tokens that are important to query and have large weights after softmax operators. Therefore, once the key tokens are included in $R_{think}^s$, the additional noise compared to $R_{think}^l$ can be suppressed to obtain the appropriate response.

This observation also inspires us to select key reasoning tokens from $R_{think}^s$ to reduce the number of transferred tokens for large models and potential noisy tokens. 
\begin{align}
&R_{think}^{'s} = \text{Select}(R_{think}^s)\\
&A_{think}^l = VLM_l(\text{Q+$R_{think}^{'s}$+Prompt$_{simple}$})
\end{align}
where $\text{Select}()$ can pick key tokens according to attention weights (\eg, $q_s^j$) from $R_{think}^s$.

We provide a detailed layer-wise analysis of the attention sparsity in Appendix~\ref{sec:sparse}, confirming that the sparsity pattern holds consistently across layers, with the top 20\% of reasoning tokens contributing approximately 80\% of the attention weight. The detailed selection strategy is demonstrated in \cref{sec:sparse_transfer}.

\begin{table*}[t]
    \centering
    \caption{Comparison of the averaged number of generated tokens with Qwen3-VL models.}
    \label{tab:qwen3_token}
    \resizebox{0.9\textwidth}{!}{
    \begin{tabular}{c|ccc|ccc|ccc|ccc|ccc|ccc}
    \toprule
    \multirow{2}{*}{Qwen3-VL} & \multicolumn{3}{c|}{POPE} & \multicolumn{3}{c|}{MMMU} & \multicolumn{3}{c|}{MMBench} & \multicolumn{3}{c|}{ChartQA} & \multicolumn{3}{c|}{$\text{InfoVQA}_{val}$} & \multicolumn{3}{c}{RealWorldQA} \\
    \cmidrule(lr){2-4} \cmidrule(lr){5-7} \cmidrule(lr){8-10} \cmidrule(lr){11-13} \cmidrule(lr){14-16} \cmidrule(lr){17-19}
    & S & E & R & S & E & R & S & E & R & S & E & R & S & E & R & S & E & R \\
    \midrule
    2B & 1.0 & 105.4 & 266.1 & 101.7 & 1448.9 & 2019.1 & 3.7 & 449.6 & 569.3 & 3.5 & 95.8 & 256.3 & 3.4 & 74.7 & 192.5 & 1.4 & 155.6 & 244.4\\
    4B & 1.0 & 97.0 & 168.6 & 47.8 & 1122.9 & 1822.2 & 2.9 & 336.1 & 475.2 & 3.6 & 276.1 & 352.2 & 3.6 & 240.7 & 288.0 & 1.1 & 209.0 & 346.2 \\
    8B & 1.0 & 34.4 & 149.3 & 45.8 & 944.6 & 1548.7 & 2.2 & 106.0 & 426.5 & 3.5 & 249.8 & 294.3 & 3.5 & 177.9 & 228.2 & 1.0 & 221.5 & 297.0 \\
    \bottomrule
    \end{tabular}
    }
\end{table*}

\cref{fig:placeholder} illustrates the framework of the proposed multi-agent inference strategy. We keep large models to output short sequences while using small models for reasoning. With reasoning transfer, we can simplify the framework by adopting $R_{think}^s$ in lieu of $A_{think}^s$ for mutual verification to further reduce the output tokens from the small model.

\section{Experiments}

In this section, we conduct experiments on standard benchmarks to demonstrate the performance and efficiency of different models. The detailed settings are elaborated as follows.

\paragraph{Evaluation Platform} We utilize LMMs-Eval~\cite{zhang2024lmmsevalrealitycheckevaluation}, a VLM evaluation framework, and vLLM~\cite{vllm} to evaluate models on H100 (96G). Unless otherwise specified, all sample parameters and task setups follow the official instructions.

\paragraph{Benchmark Tasks} To evaluate the performance of models with varying sizes extensively, we have 6 diverse tasks with different output formats in the comparison. POPE~\cite{li2023evaluating} is a yes/no benchmark that aims to evaluate hallucination. MMBench~\cite{liu2024mmbench} covers general multimodal understanding tasks in a multiple-choice format. Beside the simple output format, we have MMMU~\cite{yue2024mmmu} and RealWorldQA~\footnote{https://huggingface.co/datasets/xai-org/RealworldQA} to evaluate the comprehensive reasoning ability of models with both multi-choice and open-ended questions. Finally, two data sets with open-ended questions, \ie, ChartQA~\cite{chartqa} and InfographicVQA~\cite{infovqa}, are included for further demonstration.

\paragraph{Vision Language Model Families} Two popular open-sourced VLM families: Qwen3-VL-Instruct and InternVL3.5 are included in the comparison. Concretely, three models from Qwen3-VL, \ie, \textbf{Qwen3-VL-2/4/8B-Instruct}~\cite{Qwen3VL_GitHub_2025} and five models from InternVL3.5, \ie, \textbf{InternVL3.5-1/2/4/8/14B}~\cite{internvl} are adopted for evaluation.

\paragraph{Prompts for Output Token Control}
Previous work~\cite{guo2025deepseek} shows that the test-time scaling criterion is an effective strategy to enhance model capacity with more inference tokens. To trade off between performance and the number of output tokens, we design three different prompts for each question to control different numbers of generated tokens. Concretely, we have \textbf{Simple} (\textbf{S}) to let the model respond to the query without thinking. \textbf{Explain} (\textbf{E}) aims to encourage models to provide a simple explanation before giving the final answer. Following \cite{kojima2022large}, \textbf{Reasoning} (\textbf{R}) is applied to explicitly prompt models to think and explain with more thinking tokens. Finally, we add an answer format, \ie, \texttt{[[answer]]}, to help parse the answer from a long response. The details are as follows.
\begin{itemize}[noitemsep, topsep=0pt]
\item \textbf{Simple} (\textbf{S}): \textbf{Without thinking, directly} answer the question using a single word or phrase in the format: [[answer]]
\item \textbf{Explain} (\textbf{E}): \textbf{Provide a simple explanation}, then answer the question using a single word or phrase in the format: [[answer]]
\item \textbf{Reasoning} (\textbf{R}): \textbf{Please think step by step, provide a detailed explanation}, then answer the question using a single word or phrase in the format: [[answer]]
\end{itemize}

\subsection{Performance vs. Output Tokens} \label{sec:qwen_perf}

\paragraph{Comparison within Qwen3-VL Family} 
First, we compare the performance on Qwen 2/4/8B models. Since it is challenging to control the exact number of output tokens, instead we use different prompts and \cref{tab:qwen3_token} shows the averaged number of output tokens on different tasks. We can find that our ``Simple'' prompt can reduce the number of output tokens effectively, where models only generate no more than 10 tokens on 5 out of 6 tasks. With the ``Reasoning'' prompt, the number of output tokens increases significantly while the ``Explain'' prompt obtains an expected number of tokens that is between ``Simple'' and ``Reasoning'' in most cases. The comparison shows that our proposed prompts can adjust the number of output tokens appropriately.

\begin{table}[h]
    \centering
    \caption{Comparison of performance with Qwen3-VL models on POPE. (S2) denotes the 2-stage strategy as in Eqn.~\ref{eq:2stage}.} 
    \label{tab:qwen_perf_pope}
    \resizebox{0.8\linewidth}{!}{
    \begin{tabular}{c|ccccc}
    
    \toprule
    \multirow{2}{*}{Qwen3-VL}& \multicolumn{5}{c}{POPE}  \\
    \cmidrule(lr){2-6} 
     & S & E & E(S2) & R & R(S2)  \\
    \midrule
    2B & 88.6 & 88.8 & 89.2 & 88.4 & 89.9  \\
    4B & 88.7 & 88.0 & 89.6 & 87.8 & 88.9 \\
    8B & 87.1 & 87.3 & 87.8 & 88.2 & 89.5 \\
    \bottomrule
    \end{tabular}  
    }
\end{table}

With these prompts, we first evaluate the performance of models with different output tokens on POPE in \cref{tab:qwen_perf_pope}. 

\begin{table*}[t]
    \centering
    \caption{Comparison of performance with Qwen3-VL models. $^\dagger$ denotes original results without 2-stage strategy.} 
    \label{tab:qwen_perf}
    \resizebox{0.95\textwidth}{!}{
    \begin{tabular}{c|ccc|ccc|ccc|ccc|ccc}
    \toprule
    \multirow{2}{*}{Qwen3-VL}&   \multicolumn{3}{c|}{MMBench} &\multicolumn{3}{c|}{MMMU} &\multicolumn{3}{c|}{ChartQA} & \multicolumn{3}{c|}{InfoVQA} & \multicolumn{3}{c}{RealWorldQA} \\
    \cmidrule(lr){2-4} \cmidrule(lr){5-7} \cmidrule(lr){8-10} \cmidrule(lr){11-13} \cmidrule(lr){14-16} 
     & S & E & R & S & E$^\dagger$  & R$^\dagger$   & S & E  & R & S & E  & R &S & E  & R \\
    \midrule
    2B  &  76.8  & 75.4  & 75.0 &44.2 & 46.9   & 47.9   &77.80  & 82.72 & 83.72 & 68.64 & 71.50  & 74.57 & 60.65  & 63.79  & 64.58\\
    4B & 83.7   & 85.0  & 84.5 &48.3 & 60.0   & 61.3  & 83.12  & 83.04 & 83.16 & 79.96  & 82.87 & 83.08 & 71.24 & 70.59  & 69.93\\
    8B &  85.2  & 86.0  & 85.9 &55.6 & 62.0  & 64.3   &84.72  & 87.68 & 86.20 & 82.80 & 86.37 & 86.66 & 67.58 & 72.68 & 70.33\\
    \bottomrule
    \end{tabular}  
    }
    
\end{table*}

From the comparison, we can observe that with more output tokens, the performance for 8B model improves. However, the performance on other model sizes is not monotonic with respect to the number of output tokens. 
The primary bottleneck is an \textit{instruction-following (IF) failure}: after generating a long reasoning chain, the small model frequently forgets to output the final answer in the required format (see Example 2 in \cref{sec:append_example} for a qualitative illustration).

To disentangle genuine reasoning capability from formatting limitations, we apply a 2-stage decoding strategy (Eqn.~\ref{eq:2stage}), denoted as (S2) in \cref{tab:qwen_perf_pope}. The first stage generates raw reasoning, and the second stage forces formatted answer extraction. With appropriate KV-Cache reuse, the additional overhead is modest. This strategy confirms that models \textit{do} benefit from thinking tokens once formatting errors are eliminated. Unless otherwise noted (marked with $\dag$), all Explain (E) and Reasoning (R) results reported in this paper adopt S2, ensuring that the comparison reflects true reasoning ability rather than formatting artifacts.

Then, we report the results on remaining data sets in \cref{tab:qwen_perf} with 2-stage strategy for E and R if not specified. 
We can find that with more tokens from E and R, the 2B model still lags behind in performance on those tasks compared to 8B one with the Simple prompt, \ie 8B(S). According to the decoding profiling in \cref{tab:qwen_decoding}, the 8B model can be more efficient than 2B with reasoning, \eg, 3.5/47.26 v.s. 256.3/61.80 on ChartQA. Compared with the 2B model, the 4B(S) is worse than 8B(S) on 4 out of 5 tasks but can achieve comparable or even better performance using E/R variants with a significant overhead from extra generated tokens. The comparison demonstrates that a large model can be efficient with fewer output tokens but still provide good performance. 
Notably, the need for S2 itself highlights a practical cost of long-chain reasoning. As we show in \cref{sec:exp_mai}, our framework avoids this overhead entirely by restricting the answering model to short outputs.
The comparison on InternVL3.5 can be found in \cref{sec:append_internvl}.

\begin{takeawaybox}{}{}  
\includegraphics[width=0.8em]{fig/light-bulb.png}~\textbf{Takeaway 2:} 
Test-time scaling with small models comes with two compounding costs: smaller models with extended reasoning still underperform larger models with simple prompts at lower latency, and long output chains cause instruction-following degradation requiring an extra decoding stage. Together, these favor large models with short outputs over small models with verbose reasoning.
\end{takeawaybox}

\subsection{Reasoning Transfer} 

While large models perform well with simple prompts, more reasoning tokens still effectively improve their performance (\cref{tab:qwen_perf}). Concretely, we reuse the cheap reasoning tokens $R_{think}^s$ generated by the small model and prepend them as the contextual prompt for the large model (Eqn.~\ref{eq:rt}), which then directly answers the query using the Simple prompt.

A natural concern arises: \textit{Will the incorrect or noisy reasoning steps from the small model mislead the large model?} Fortunately, as mathematically grounded in Corollary~\ref{cor:1}, the Transformer's self-attention mechanism inherently acts as a noise filter. It concentrates attention weights on a few key semantic signals while bounding the norm difference caused by irrelevant or noisy tokens. Therefore, as long as the small model captures some critical visual or semantic cues, the large model can selectively attend to them while suppressing the flawed reasoning paths.

We conduct empirical comparisons using the Qwen models, reporting the performance in \cref{tab:2b_to_all}.
\newcommand{\numupval}[2]{\textbf{#1}\rlap{~{\tiny\textcolor{green}{\textbf{#2}}}}}
\newcommand{\numup}[2]{\textbf{#1}\rlap{~{\tiny\textcolor{green}{\textbf{#2}}}}}
\newcommand{\numdown}[2]{#1\rlap{~{\tiny\textcolor{red}{\textbf{#2}}}}}
\newcommand{\gray}[1]{\textcolor{gray}{#1}}

\begin{table}[h]
    \centering
    \caption{Comparison of reasoning transfer that uses reasoning tokens from the 2B/4B model to replace the reasoning tokens for 4B/8B models with the 2-stage strategy as in Eqn.~\ref{eq:2stage}.
    Percentage changes show improvement (\textcolor{green}{green}) or degradation (\textcolor{red}{red}) compared to baseline with the ``Simple'' prompt, \ie, ``-S''.}
    \label{tab:2b_to_all}
    \setlength{\tabcolsep}{3pt}
    \resizebox{\linewidth}{!}{
    \begin{tabular}{lccccccc}
            \toprule
            Qwen3-VL  & POPE & MMMU & MMBench & ChartQA & InfoVQA & RealWorldQA\\
            \midrule
            2B-R & 88.4 & 47.9 & 72.8 & 72.2 & 71.4 & 60.5  \\
            \midrule
            4B-S & 88.7 & 48.3 & 83.7 & 83.1 & 80.0 & 71.2 \\
            \midrule
            4B-R & 88.9 & 61.3 & 84.5 & 83.2 & 83.1 & 69.9 \\
            4B + 2B-R & \numupval{91.0}{+2.6\%} & \numup{55.9}{+15.7\%} & \numup{84.1}{+0.5\%} & \numdown{82.4}{-0.8\%} & \numdown{79.7}{-0.4\%} & \numup{71.6}{+0.6\%} \\
            \midrule
            8B-S & 87.1 & 55.6 & 85.2 & 84.7 & 82.8 & 67.6\\
            \midrule
            8B-R & 89.5 & 64.3 & 85.9 & 86.2 & 86.7 & 70.3\\
            8B + 2B-R & \numupval{89.2}{+2.4\%} & \numup{62.9}{+13.1\%} & \numdown{85.1}{-0.1\%}  & \numup{85.3}{+0.7\%} & \numup{82.8}{+0.0\%} & \numup{68.9}{+1.9\%} \\
            8B + 4B-R & \numup{87.9}{+0.9\%} & \numup{62.9}{+13.1\%} & \numdown{85.1}{-0.1\%} & \numup{86.0}{+1.5\%} & \numup{84.1}{+1.6\%} & \numup{70.2}{+3.8\%} \\
            \bottomrule
    \end{tabular}
    }
\end{table}

The results strongly validate our theoretical analysis. Although the original reasoning tokens from the 2B model (2B-R) yield worse standalone performance than 4B-S and 8B-S, using them as the prompt for the 8B model improves the baseline performance substantially (e.g., clear improvements on 4 out of 6 tasks compared to 8B-S). More importantly, any observed degradation is exceptionally mild, confirming that the large model successfully tolerates the noisy reasoning tokens and extracts useful signals. Note that reasoning transfer inherently avoids the instruction-following degradation discussed in \cref{sec:qwen_perf}: since the large model only generates a short answer via the Simple prompt, no 2-stage correction is needed.

\subsection{Sparse Reasoning Transfer} \label{sec:sparse_transfer}

To further reduce the number of input tokens for large models and potential noise in reasoning tokens from small models, we explore whether the large model strictly requires a continuous reasoning chain. We empirically observe that transferring only a sparse subset of key tokens with large attention weights is sufficient for logical understanding. This aligns with Corollary~\ref{cor:1}, which indicates that preserving a fraction of dominating tokens can effectively bound the representation shift.

Concretely, reasoning tokens from small models will be ranked in descending order according to the attention weights. Then, a subset of top tokens with the total weights up to 80\% of the original weights, \ie, $\delta=0.8$ in  Corollary~\ref{cor:1}, is sampled for large models.

   \begin{table}[h]
    \centering
    \caption{Comparison of attention weights from different layers (L\#) for sparse reasoning token transfer (8B+2B-R) performance on POPE. Tokens are sampled according to the attention weights and the total weight of sampled tokens is 80\% of the original set. \#T denotes the number of selected tokens.}
    \label{tab:sparsity_results}
    
    \resizebox{\linewidth}{!}{
    \begin{tabular}{l|c|cccccccc|c}
        \toprule
        \multirow{2}{*}{Config}& \multirow{2}{*}{8B-S}&\multicolumn{8}{c|}{Sparse Reasoning Transfer}&\multirow{2}{*}{Full} \\
        \cmidrule(lr){3-10}
        &&L4&L8&L10&L12&L14&L16&L22&L24&\\
        \midrule
        
        Avg. \#T & - & 145.9&153.6&85.9&90.5&89.0&121.1&112.1&209.5&266.1\\
        \midrule
        F1& 87.1&89.3&89.0&88.7&89.2&88.9&89.5&89.6&89.6&89.2\\
        \bottomrule
    \end{tabular}
    }
\end{table}

\begin{table*}[t]
    \centering
    \caption{Multi-Agent Inference based on Qwen3-VL-2B/8B. ``MV'' denotes the mutual verification and ``RT'' is for reasoning transfer. The best result is in bold.}
    \label{tab:qwen_2/8B}
    \resizebox{\linewidth}{!}{
    \begin{tabular}{l|c|ccc|c|ccc|c}
        \toprule
        \multirow{2}{*}{Dataset}&\multirow{2}{*}{\#Test}&\multicolumn{4}{c|}{Multi-Agent Inference (MAI)}  &\multicolumn{3}{c|}{Baseline} &\multicolumn{1}{c}{Full RT} \\
        \cmidrule(lr){3-6} \cmidrule(lr){7-9} \cmidrule(lr){10-10}
         &  & MV$_{2B(S)/8B(S)}$ & MV$_{2B(R)/8B(S)}$ & RT & Final Perf. & 2B(S) & 2B(R) & 8B(S) & 8B+2B-R \\
        \midrule
           POPE      & 9000  & 8507 (94.52\%) & 46 (0.51\%)    & 447 (4.97\%)  & \textbf{89.4} & 88.6 & 88.4 & 87.1 & 89.2 \\
           MMMU      & 900   & 498 (55.33\%)  & 142 (15.78\%)  & 260 (28.89\%) & 62.0 & 44.2 & 47.9 & 55.6 & \textbf{62.9} \\
           MMBench   & 4329  & 3736 (86.34\%) & 262 (6.05\%) & 330 (7.62\%) & \textbf{85.5} & 76.8 & 72.8 & 85.2 & 85.1 \\
           ChartQA   & 2500  & 1800 (72.00\%) & 195 (7.80\%)   & 505 (20.20\%) & \textbf{86.8} & 77.8 & 72.2 & 84.7 & 85.3 \\
           InfoVQA   & 2801  & 1728 (61.69\%) & 271 (9.68\%)   & 802 (28.63\%) & \textbf{82.9} & 68.6 & 71.4 & 82.8 & 82.8 \\
           RealWorldQA & 765  & 509 (66.54\%) & 105 (13.73\%) & 151 (19.74\%) & 68.4 & 60.7 & 60.5 & 67.6 & \textbf{68.9} \\
        \bottomrule
    \end{tabular}
    }

\end{table*}

\begin{table}[t]
    \centering
    \caption{Comparison of total inference time (s) on benchmarks.}
    \label{tab:qwen_multiagent_latency}
    \resizebox{\linewidth}{!}{
    \begin{tabular}{l|cccccc}
        \toprule
        {Method} & {POPE} & {MMBench} & {MMMU} & {ChartQA} & {InfoVQA} & {RealWorldQA}\\  
        \midrule

        Full RT (s) & 39,865 & 40,718 & 43,924 & 10,805 & 9,201  & 561\\ 
        \midrule
        {MAI (s)} & \textbf{4,954} & \textbf{8,676} & \textbf{25,025} & \textbf{4,158} & \textbf{5,094} & \textbf{268}\\ 
        {Speedup}     & \textbf{8.05$\times$} & \textbf{4.69$\times$} & \textbf{1.76$\times$} & \textbf{2.60$\times$} & \textbf{1.81$\times$} & \textbf{2.09$\times$}\\ 
        \bottomrule
    \end{tabular}
    }
\end{table}

Considering that there are multiple layers containing attention maps in transformers, we conduct the ablation on different layers and summarize the results on POPE in \cref{tab:sparsity_results}. 

Compared to the original reasoning transfer (\ie, ``Full''), our sparse reasoning transfer strategy achieves the comparable or even better performance with significantly fewer reasoning tokens, which confirms our theoretical analysis. In addition, different layers demonstrate different distributions of attention scores. We observe a convex curve in token usage, where layers L10-L14 balance the number of selected tokens and performance well. Finally, all transferred reasoning tokens can help improve the performance over the baseline 8B-S, which demonstrates the effectiveness of reasoning transfer. More detailed layer-wise illustration can be found in Sec.~\ref{sec:sparse}.

\subsection{Multi-Agent Inference} \label{sec:exp_mai}
Finally, we evaluate the proposed multi-agent framework that leverages mutual verification and reasoning transfer to minimize the number of output tokens from large models. The comparison is mainly based on Qwen models. We consider the agent system consists of 2B-8B models. The detailed results are shown in \cref{tab:qwen_2/8B}.

First, compared with the strong baseline, \ie 8B(S), the multi-agent inference can improve the performance consistently. It implies that with the help of the small model, the large model can leverage the transferred tokens effectively for challenging queries. Second, mutual verification (MV) can reduce the number of expensive model calls, \eg 2B(R) and reasoning transfer (RT) for the 8B model, significantly. For example, at least 55\% of inferences achieve agreement between large and small models with the simple prompt over different tasks. There are up to 15\% test examples that can obtain the same results after reasoning by small models. Less than 30\% of queries will call large models with reasoning transfer. 

Remarkably, compared with the full reasoning transfer, our multi-agent system shows a comparable or even better performance (Full RT, \eg, 86.8\% vs. 85.3\% on ChartQA), which confirms the effectiveness of mutual verification.

\begin{takeawaybox}{}{}  
\includegraphics[width=0.8em]{fig/light-bulb.png}~\textbf{Takeaway 3:} 
 Reasoning tokens from small models can be transferred to large models as input context, converting expensive sequential decoding into cheap prefilling. Despite containing noisy or incorrect reasoning steps, the large model's self-attention selectively attends to key semantic signals while suppressing flawed paths (Corollary~\ref{cor:1}), yielding consistent improvements over the no-reasoning baseline.
\end{takeawaybox}

\paragraph{Latency Estimation: A Conservative Upper Bound.}
To quantify end-to-end efficiency independent of specific serving frameworks or hardware optimizations, we formulate a strictly conservative upper bound for the total inference time. We estimate this by summing the sequential execution costs without assuming any parallelization. The detailed formulation is provided in Appendix~\ref{sec:append_latency_calculation}. Based on this estimation, we report the total inference time for the entire dataset in \cref{tab:qwen_multiagent_latency}. Even under this pessimistic worst-case scenario, our MAI framework significantly accelerates inference over full reasoning transfer, achieving speedups ranging from $1.76\times$ (MMMU) to $8.05\times$ (POPE). Note that this theoretical bound assumes fully sequential execution, in practice, continuous batching and parallel execution of Stage 1 would further reduce this latency.

\subsection{Comparison with Speculative Decoding}
\label{sec:sd_comparison}

To assess token-level speculative decoding (SD), we tested a 2B-draft/8B-target setup on ChartQA via HuggingFace. Empirically, SD is 16\% slower than standard decoding (\cref{tab:sd_comparison}) due to high draft rejection (~40\%) and severe token-level context-switching overheads (Appendix~\ref{sec:append_latency_impl_details}).

In contrast, MAI transfers complete reasoning contexts to the response level. Requiring only a single 8B invocation, MAI is natively compatible with optimized engines such as vLLM, achieving a 2.40$\times$ speedup over the 8B-R baseline while boosting accuracy to 86.52\%. Extended discussions on memory and scalability are provided in Appendix~\ref{sec:append_discussion}.

\begin{table}[htbp]
\centering
\caption{Comparison of decoding strategies on ChartQA. $^\dag$HF latency excludes model loading, whereas vLLM is end-to-end. All reported latencies exclude S2 overhead, which is required by baselines for best accuracy but avoided by MAI.}
\label{tab:sd_comparison}
\resizebox{\linewidth}{!}{%
\begin{tabular}{l l c c cc c cc}
\toprule
\multirow{2}{*}{\textbf{Method}} & \multirow{2}{*}{\textbf{Backend}} & \multirow{2}{*}{\textbf{BS}} & \multirow{2}{*}{\makecell{\textbf{Accept}\\\textbf{Rate}}} & \multirow{2}{*}{\makecell{\textbf{Total$^\dag$}\\(s) $\downarrow$}} & \multirow{2}{*}{\makecell{\textbf{Latency$^\dag$}\\(s/sample) $\downarrow$}} & \multirow{2}{*}{\textbf{Speedup}} & \multicolumn{2}{c}{\textbf{ChartQA Acc.\ (\%)}} \\
\cmidrule(lr){8-9}
 & & & & & & & w/o S2 & w/ S2 \\
\midrule
8B-R            & HF   & 1  & N/A     & 27,195 & 10.88 & 1.00$\times$ & 77.08 & 85.52 \\
SD (8B+2B)      & HF   & 1  & 60.39\% & 31,540 & 12.62 & 0.86$\times$ & 77.24 & 85.40 \\
\midrule
8B-R            & vLLM & 32 & N/A     & 1,122  & 0.449 & 1.00$\times$ & 77.68 & 86.32 \\
MAI (8B+2B)     & vLLM & 32 & N/A     & 468    & 0.187 & 2.40$\times$ & \textbf{86.52} & - \\
\bottomrule
\end{tabular}%
}
\end{table}

\section{Conclusion}
\label{sec:conclusion}

In this work, we revisit the often-overlooked role of output token length in determining the end-to-end inference efficiency of VLMs. While prior research has primarily focused on optimizing model size and visual token reduction, our profiling analysis demonstrates that the number of generated tokens is a critical factor influencing latency and cost.

With a comprehensive evaluation of SoTA VLMs across 6 real-world benchmarks, we observe that larger models achieve strong performance with substantially fewer output tokens, whereas smaller models require longer, more verbose reasoning chains to reach comparable accuracy. Therefore, large models can be more efficient than small models in terms of achieving a targeted performance. According to this observation, we developed a multi-agent inference framework with mutual verification and reasoning transfer to leverage large models to help the accurate and efficient inference.

Currently, we have not included the GPU memory as a constraint for inference. Exploring those additional constraints and extending our multi-agent framework to save memories can be our future work.

\bibliography{main}

@String(CVPR= {IEEE Conf. Comput. Vis. Pattern Recog.})

@String(ECCV= {Eur. Conf. Comput. Vis.})

@String(CVPR  = {CVPR})

@String(ECCV  = {ECCV})

@inproceedings{WilhelmWK25,
  author       = {Patrick Wilhelm and
                  Thorsten Wittkopp and
                  Odej Kao},
  editor       = {Eiko Yoneki and
                  Amir H. Payberah},
  title        = {Beyond Test-Time Compute Strategies: Advocating Energy-per-Token in
                  {LLM} Inference},
  booktitle    = {Workshop on Machine Learning and Systems, EuroMLSys},
  pages        = {208--215},
  publisher    = {{ACM}},
  year         = {2025}
}

@article{internvl,
  author       = {Weiyun Wang and
                  Zhangwei Gao and
                  Lixin Gu and
                  Hengjun Pu and
                  Long Cui and
                  Xingguang Wei and
                  Zhaoyang Liu and
                  Linglin Jing and
                  Shenglong Ye and
                  Jie Shao and
                  Zhaokai Wang and
                  Zhe Chen and
                  Hongjie Zhang and
                  Ganlin Yang and
                  Haomin Wang and
                  Qi Wei and
                  Jinhui Yin and
                  Wenhao Li and
                  Erfei Cui and
                  Guanzhou Chen and
                  Zichen Ding and
                  Changyao Tian and
                  Zhenyu Wu and
                  JingJing Xie and
                  Zehao Li and
                  Bowen Yang and
                  Yuchen Duan and
                  Xuehui Wang and
                  Zhi Hou and
                  Haoran Hao and
                  Tianyi Zhang and
                  Songze Li and
                  Xiangyu Zhao and
                  Haodong Duan and
                  Nianchen Deng and
                  Bin Fu and
                  Yinan He and
                  Yi Wang and
                  Conghui He and
                  Botian Shi and
                  Junjun He and
                  Yingtong Xiong and
                  Han Lv and
                  Lijun Wu and
                  Wenqi Shao and
                  Kaipeng Zhang and
                  Huipeng Deng and
                  Biqing Qi and
                  Jiaye Ge and
                  Qipeng Guo and
                  Wenwei Zhang and
                  Songyang Zhang and
                  Maosong Cao and
                  Junyao Lin and
                  Kexian Tang and
                  Jianfei Gao and
                  Haian Huang and
                  Yuzhe Gu and
                  Chengqi Lyu and
                  Huanze Tang and
                  Rui Wang and
                  Haijun Lv and
                  Wanli Ouyang and
                  Limin Wang and
                  Min Dou and
                  Xizhou Zhu and
                  Tong Lu and
                  Dahua Lin and
                  Jifeng Dai and
                  Weijie Su and
                  Bowen Zhou and
                  Kai Chen and
                  Yu Qiao and
                  Wenhai Wang and
                  Gen Luo},
  title        = {InternVL3.5: Advancing Open-Source Multimodal Models in Versatility,
                  Reasoning, and Efficiency},
  journal      = {CoRR},
  volume       = {abs/2508.18265},
  year         = {2025}
}

@article{smolvlm,
  title={SmolVLM: Redefining small and efficient multimodal models}, 
  author={Andrés Marafioti and Orr Zohar and Miquel Farré and Merve Noyan and Elie Bakouch and Pedro Cuenca and Cyril Zakka and Loubna Ben Allal and Anton Lozhkov and Nouamane Tazi and Vaibhav Srivastav and Joshua Lochner and Hugo Larcher and Mathieu Morlon and Lewis Tunstall and Leandro von Werra and Thomas Wolf},
  journal={arXiv preprint arXiv:2504.05299},
  year={2025}
}

@article{qwen,
  author       = {Shuai Bai and
                  Keqin Chen and
                  Xuejing Liu and
                  Jialin Wang and
                  Wenbin Ge and
                  Sibo Song and
                  Kai Dang and
                  Peng Wang and
                  Shijie Wang and
                  Jun Tang and
                  Humen Zhong and
                  Yuanzhi Zhu and
                  Ming{-}Hsuan Yang and
                  Zhaohai Li and
                  Jianqiang Wan and
                  Pengfei Wang and
                  Wei Ding and
                  Zheren Fu and
                  Yiheng Xu and
                  Jiabo Ye and
                  Xi Zhang and
                  Tianbao Xie and
                  Zesen Cheng and
                  Hang Zhang and
                  Zhibo Yang and
                  Haiyang Xu and
                  Junyang Lin},
  title        = {Qwen2.5-VL Technical Report},
  journal      = {CoRR},
  volume       = {abs/2502.13923},
  year         = {2025}
}

@inproceedings{cot,
  author       = {Jason Wei and
                  Xuezhi Wang and
                  Dale Schuurmans and
                  Maarten Bosma and
                  Brian Ichter and
                  Fei Xia and
                  Ed H. Chi and
                  Quoc V. Le and
                  Denny Zhou},
  editor       = {Sanmi Koyejo and
                  S. Mohamed and
                  A. Agarwal and
                  Danielle Belgrave and
                  K. Cho and
                  A. Oh},
  title        = {Chain-of-Thought Prompting Elicits Reasoning in Large Language Models},
  booktitle    = {NeurIPS},
  year         = {2022}
}

@article{mmtok,
  author       = {Sixun Dong and
                  Juhua Hu and
                  Mian Zhang and
                  Ming Yin and
                  Yanjie Fu and
                  Qi Qian},
  title        = {MMTok: Multimodal Coverage Maximization for Efficient Inference of
                  VLMs},
  journal      = {CoRR},
  volume       = {abs/2508.18264},
  year         = {2025}
}

@inproceedings{liu2024mmbench,
  title={Mmbench: Is your multi-modal model an all-around player?},
  author={Liu, Yuan and Duan, Haodong and Zhang, Yuanhan and Li, Bo and Zhang, Songyang and Zhao, Wangbo and Yuan, Yike and Wang, Jiaqi and He, Conghui and Liu, Ziwei and others},
  booktitle={ECCV},
  pages={216--233},
  year={2024},
  organization={Springer}
}

@inproceedings{yue2024mmmu,
  title={Mmmu: A massive multi-discipline multimodal understanding and reasoning benchmark for expert agi},
  author={Yue, Xiang and Ni, Yuansheng and Zhang, Kai and Zheng, Tianyu and Liu, Ruoqi and Zhang, Ge and Stevens, Samuel and Jiang, Dongfu and Ren, Weiming and Sun, Yuxuan and others},
  booktitle={CVPR},
  pages={9556--9567},
  year={2024}
}

@article{li2023evaluating,
  title={Evaluating object hallucination in large vision-language models},
  author={Li, Yifan and Du, Yifan and Zhou, Kun and Wang, Jinpeng and Zhao, Wayne Xin and Wen, Ji-Rong},
  journal={arXiv:2305.10355},
  year={2023}
}

@book{convex,
  author       = {Stephen P. Boyd and
                  Lieven Vandenberghe},
  title        = {Convex Optimization},
  publisher    = {Cambridge University Press},
  year         = {2014}
}

@article{guo2025deepseek,
  title={Deepseek-r1: Incentivizing reasoning capability in llms via reinforcement learning},
  author={Guo, Daya and Yang, Dejian and Zhang, Haowei and Song, Junxiao and Zhang, Ruoyu and Xu, Runxin and Zhu, Qihao and Ma, Shirong and Wang, Peiyi and Bi, Xiao and others},
  journal={arXiv preprint arXiv:2501.12948},
  year={2025}
}

@misc{zhang2024lmmsevalrealitycheckevaluation,
      title={LMMs-Eval: Reality Check on the Evaluation of Large Multimodal Models}, 
      author={Kaichen Zhang and Bo Li and Peiyuan Zhang and Fanyi Pu and Joshua Adrian Cahyono and Kairui Hu and Shuai Liu and Yuanhan Zhang and Jingkang Yang and Chunyuan Li and Ziwei Liu},
      year={2024},
      eprint={2407.12772},
      archivePrefix={arXiv},
      primaryClass={cs.CL} 
}

@article{kojima2022large,
  title={Large language models are zero-shot reasoners},
  author={Kojima, Takeshi and Gu, Shixiang Shane and Reid, Machel and Matsuo, Yutaka and Iwasawa, Yusuke},
  journal={NeruIPS},
  volume={35},
  pages={22199--22213},
  year={2022}
}

@software{Qwen3VL_GitHub_2025,
  author       = {{Qwen Team}},
  title        = {Qwen3-VL: Multimodal Vision-Language Model Series},
  organization = {Alibaba Cloud},
  year         = {2025}
}

@inproceedings{vllm,
  title={Efficient Memory Management for Large Language Model Serving with PagedAttention},
  author={Woosuk Kwon and Zhuohan Li and Siyuan Zhuang and Ying Sheng and Lianmin Zheng and Cody Hao Yu and Joseph E. Gonzalez and Hao Zhang and Ion Stoica},
  booktitle={Proceedings of the ACM SIGOPS 29th Symposium on Operating Systems Principles},
  year={2023}
}

@inproceedings{llava,
  author       = {Haotian Liu and
                  Chunyuan Li and
                  Qingyang Wu and
                  Yong Jae Lee},
  editor       = {Alice Oh and
                  Tristan Naumann and
                  Amir Globerson and
                  Kate Saenko and
                  Moritz Hardt and
                  Sergey Levine},
  title        = {Visual Instruction Tuning},
  booktitle    = {NeurIPS},
  year         = {2023}
}

@inproceedings{visionzip,
  author       = {Senqiao Yang and
                  Yukang Chen and
                  Zhuotao Tian and
                  Chengyao Wang and
                  Jingyao Li and
                  Bei Yu and
                  Jiaya Jia},
  title        = {VisionZip: Longer is Better but Not Necessary in Vision Language Models},
  booktitle    = {CVPR},
  pages        = {19792--19802},
  publisher    = {Computer Vision Foundation / {IEEE}},
  year         = {2025}
}

@article{offloaded,
  title={Offloaded Reasoning: Efficient Inference for Large Language Models via Modular Reasoning and Refinement},
  author={Jindala, Ishan and Tanejaa, Jayant and Badrinathb, Chandana and Kapura, Vikas and Sharmaa, Sachin Dev}
}

@article{scot,
  author       = {Jikai Wang and
                  Juntao Li and
                  Lijun Wu and
                  Min Zhang},
  title        = {Efficient Reasoning for LLMs through Speculative Chain-of-Thought},
  journal      = {CoRR},
  volume       = {abs/2504.19095},
  year         = {2025}
}

@article{liu2025small,
  title={Small drafts, big verdict: Information-intensive visual reasoning via speculation},
  author={Liu, Yuhan and Qin, Lianhui and Wang, Shengjie},
  journal={arXiv preprint arXiv:2510.20812},
  year={2025}
}

@inproceedings{chartqa,
  title={Chartqa: A benchmark for question answering about charts with visual and logical reasoning},
  author={Masry, Ahmed and Do, Xuan Long and Tan, Jia Qing and Joty, Shafiq and Hoque, Enamul},
  booktitle={Findings of the association for computational linguistics: ACL 2022},
  pages={2263--2279},
  year={2022}
}

@inproceedings{infovqa,
  title={Infographicvqa},
  author={Mathew, Minesh and Bagal, Viraj and Tito, Rub{\`e}n and Karatzas, Dimosthenis and Valveny, Ernest and Jawahar, CV},
  booktitle={Proceedings of the IEEE/CVF Winter Conference on Applications of Computer Vision},
  pages={1697--1706},
  year={2022}
}

@inproceedings{leviathan2023fast,
  title={Fast inference from transformers via speculative decoding},
  author={Leviathan, Yaniv and Kalman, Matan and Matias, Yossi},
  booktitle={International Conference on Machine Learning},
  pages={19274--19286},
  year={2023},
  organization={PMLR}
}

@article{liu2025thought,
  title={Thought manipulation: External thought can be efficient for large reasoning models},
  author={Liu, Yule and Zheng, Jingyi and Sun, Zhen and Peng, Zifan and Dong, Wenhan and Sha, Zeyang and Cui, Shiwen and Wang, Weiqiang and He, Xinlei},
  journal={arXiv preprint arXiv:2504.13626},
  year={2025}
}

@article{r2r,
  title={R2R: Efficiently Navigating Divergent Reasoning Paths with Small-Large Model Token Routing},
  author={Fu, Tianyu and Ge, Yi and You, Yichen and Liu, Enshu and Yuan, Zhihang and Dai, Guohao and Yan, Shengen and Yang, Huazhong and Wang, Yu},
  journal={arXiv preprint arXiv:2505.21600},
  year={2025}
}

@article{specreason,
  title={SpecReason: Fast and Accurate Inference-Time Compute via Speculative Reasoning},
  author={Li, Yiming and others},
  journal={arXiv preprint arXiv:2504.07891},
  year={2025}
}

@article{speccot,
  title={SpecCoT: Accelerating Chain-of-Thought Reasoning through Speculative Exploration},
  author={Zhang, Jintian and others},
  journal={arXiv preprint arXiv:2505.12597},
  year={2025}
}
\bibliographystyle{icml2026}
\clearpage
\appendix

\section{Theoretical Analysis} \label{sec:append_proof}

\subsection{Proposition~1}
\begin{prop}\label{prop:1}
Let $\p\in\R^{d}$ be the prompt token, with attention to the reasoning tokens $X_l$ and $X_s$ as
\begin{eqnarray}
&&\p_l = \texttt{softmax}(\p X_l^\top/\lambda)X_l\\
&&\p_s = \texttt{softmax}(\p X_s^\top/\lambda)X_s
\end{eqnarray}
we have
\begin{align}
&\p_l = \sum_i^n q_i X_l^i;\ q_i = \arg\max_{q\in\Delta} (q_i X_l^i)^\top \p +\lambda H(q)\\
&\p_s = \sum_i^n q_i X_s^i;\ q_i = \arg\max_{q\in\Delta} (q_i X_s^i)^\top \p +\lambda H(q)
\end{align}
where $\Delta$ is the simplex as $\Delta = \{q:\forall i\ q_i\geq 0, \sum_i q_i=1\}$.
\end{prop}

\begin{proof}
The equivalence is from the K.K.T. condition applied for $q$ directly~\cite{convex}. 
\end{proof}

Proposition~\ref{prop:1} shows that softmax approximates the max operation with smoothness. Compared with one-hot label from max operation, it will spread weights to a few tokens mainly with positive correlations, but is still sparse with appropriate $\lambda$.

\subsection{Proof of Corollary~1}
\begin{proof}
\begin{align}\label{eq:appr}
&\|\p_l - \p_s\|_2  = \|\p_l-\p'_l+\p'_l - \p'_s+\p'_s - \p_s\|_2\\
&\leq \|\p'_l - \p'_s\|_2 + \|\p_l-\p'_l\|_2 + \|\p'_s - \p_s\|_2
\end{align}
According to Proposition 1, we have
\begin{eqnarray}
\p_l = \sum_i^n q_i X_l^i; \quad \p'_l = \sum_j^k q'_j X_l^{'j}
\end{eqnarray}
By arranging the index of tokens, we can keep the selected tokens at top $k$ such that
\begin{eqnarray}
\|\p_l-\p'_l\|_2 = \|\sum_{j}^k(q_j - q'_j)X_l^j +\sum_{j=k+1}^n q_j X_l^j\|_2\\
\leq \|\sum_{j}^k(q_j - q'_j)X_l^j\|_2 + \|\sum_{j=k+1}^n q_j X_l^j\|_2
\end{eqnarray}
According to the assumption that $\sum_{j}^k q_j\geq \delta$, we have
\begin{eqnarray}\label{eq:rest}
\|\sum_{j=k+1}^n q_j X_l^j\|_2\leq c\sum_{j=k+1}^n q_j\leq (1-\delta)c
\end{eqnarray}
For the selected tokens, since the similarity score, i.e., $\p^\top X^i$, does not change, softmax operator only amplifies the weight to make it sum to 1: $\forall j\leq k\quad q'_j = q_j/\delta$. So we have
\begin{eqnarray}\label{eq:select}
\|\sum_{j}^k(q_j - q'_j)X_l^j\|_2\leq (1-\delta)c
\end{eqnarray}
Combining Eqns.~\ref{eq:rest} and \ref{eq:select}, we have
\begin{eqnarray}
\|\p_l-\p'_l\|_2\leq 2(1-\delta)c
\end{eqnarray}
With the similar analysis, we have
\begin{eqnarray}
\|\p_s-\p'_s\|_2\leq 2(1-\delta)c
\end{eqnarray}
Taking them back to Eqn.~\ref{eq:appr} obtains the final result.
\end{proof}

\section{Experiments}

\begin{table*}[tbp]
    \centering
    \caption{Comparison of the averaged number of generated tokens with InternVL3.5 models.}
    \label{tab:internvl_token}
    \small
    \resizebox{\textwidth}{!}{
    \begin{tabular}{c*{18}{c}}
    \toprule
    \multirow{2}{*}{InternVL3.5}& \multicolumn{3}{c}{POPE} & \multicolumn{3}{c}{MMMU} & \multicolumn{3}{c}{MMBench} & \multicolumn{3}{c}{ChartQA} & \multicolumn{3}{c}{InfoVQA} & \multicolumn{3}{c}{RealWorldQA} \\
    \cmidrule(lr){2-4} \cmidrule(lr){5-7} \cmidrule(lr){8-10} \cmidrule(lr){11-13} \cmidrule(lr){14-16} \cmidrule(lr){17-19}
     & S & E & R & S & E & R & S & E & R & S & E & R & S & E & R & S & E & R \\
    \midrule
    1B  & 1.1 & 26.0 & 39.7 & 2.5 & 302.2 & 425.8 & 3.9 & 31.1 & 82.3 & 3.5 & 37.9 & 66.2 & 3.5 & 57.8 & 42.5 & 1.4 & 36.4 & 69.5 \\
    2B  & 1.0 & 12.2 & 45.2 & 262.2 & 389.9 & 508.8 & 14.0 & 66.6 & 163.1 & 3.5 & 121.4 & 143.3 & 4.4 & 78.2 & 137.6 & 1.8 & 67.6 & 122.5 \\
    4B  & 1.0 & 23.3 & 57.1 & 5.9 & 358.9 & 491.7 & 2.3 & 55.1 & 154.7 & 3.6 & 173.4 & 200.1 & 3.9 & 152.6 & 181.6 & 1.3 & 70.0 & 103.9 \\
    8B  & 1.1 & 30.5 & 46.1 & 10.6 & 262.0 & 385.5 & 2.2 & 49.5 & 141.7 & 3.9 & 120.3 & 158.2 & 3.5 & 108.6 & 142.0 & 1.6 & 54.3 & 75.7 \\
    14B & 1.4 & 19.3 & 47.0 & 11.2 & 419.0 & 484.1 & 2.5 & 46.7 & 161.0 & 3.6 & 126.7 & 147.7 & 3.4 & 52.5 & 117.6 & 3.8 & 60.8 & 71.9 \\
    \bottomrule
    \end{tabular}
    }
\end{table*}

\begin{table*}[t]
    \centering
    \caption{Comparison of performance with InternVL3.5 models. $\dagger$ denotes original results without 2-stage strategy.}
    \label{tab:internvl_perf_s2}
    \resizebox{\textwidth}{!}{
    \begin{tabular}{c|ccc|ccc|ccc|ccc|ccc|ccc}
        \toprule
        \multirow{2}{*}{InternVL} & \multicolumn{3}{c|}{POPE} & \multicolumn{3}{c|}{MMMU} & \multicolumn{3}{c|}{MMBench} & \multicolumn{3}{c|}{ChartQA} & \multicolumn{3}{c|}{InfoVQA} & \multicolumn{3}{c}{RealWorldQA} \\
        \cmidrule(lr){2-4} \cmidrule(lr){5-7} \cmidrule(lr){8-10} \cmidrule(lr){11-13} \cmidrule(lr){14-16} \cmidrule(lr){17-19}
        & S & E & R & S & E$\dagger$ & R$\dagger$ & S & E & R & S & E & R & S & E & R & S & E & R \\
        \midrule
        1B  & 83.8 & 85.6 & 83.0 & 39.3 & 43.0 & 41.6 & 68.2 & 67.8 & 68.1 & 70.7 & 65.5 & 67.2 & 56.1 & 51.9 & 51.7 & 51.9 & 58.3 & 56.1 \\
        2B  & 88.9 & 89.3 & 90.7 & 51.2 & 52.0 & 54.3 & 75.7 & 78.4 & 77.1 & 80.9 & 82.7 & 85.5 & 41.4 & 65.5 & 65.8 & 50.2 & 55.4 & 48.8 \\
        4B  & 88.8 & 93.1 & 91.7 & 55.8 & 60.4 & 59.4 & 76.5 & 81.7 & 82.5 & 83.7 & 88.6 & 87.8 & 68.9 & 73.6 & 73.9 & 59.5 & 60.7 & 42.6 \\
        8B  & 88.6 & 90.7 & 88.5 & 56.8 & 60.9 & 60.2 & 74.7 & 83.0 & 83.5 & 77.2 & 83.9 & 83.8 & 57.4 & 74.7 & 75.4 & 59.9 & 63.8 & 58.8 \\
        14B & 85.8 & 86.5 & 88.3 & 58.2 & 62.2 & 61.6 & 82.8 & 84.0 & 83.7 & 87.3 & 88.8 & 88.4 & 69.6 & 74.1 & 77.1 & 52.8 & 67.7 & 67.8 \\
        \bottomrule
    \end{tabular}
    }
\end{table*}

\subsection{Comparison within InternVL3.5 Family} \label{sec:append_internvl}
Besides Qwen family, we conduct a similar comparison with InternVL3.5 models. The number of output tokens and the performance are summarized in \cref{tab:internvl_token} and \cref{tab:internvl_perf_s2}, respectively.

First, \cref{tab:internvl_token} shows that our prompts are still effective in adjusting the number of output tokens for InternVL models. With the appropriate prompts, we can compare models with different output tokens in \cref{tab:internvl_perf_s2}. The phenomenon is similar to that on Qwen3-VL models. 4B(S)/8B(S)/14B(S) can match or even outperform smaller models with more reasoning tokens, \eg, 1B(R), 2B(R), \etc. For example, on MMMU, the 4B(S) model with about 6 tokens outperforms the 1B(E) and 2B(R) models, even when they have approximately 500 thinking tokens. It confirms our observation that large models that require fewer output tokens can be more efficient than small models to achieve a targeted performance.

\subsection{Calculation of Theoretical Latency Upper Bound}
\label{sec:append_latency_calculation}

As discussed in Section 5.4, to quantify end-to-end efficiency independent of specific serving frameworks or hardware optimizations, we formulate a strictly conservative upper bound for the total inference time. We estimate this by summing the sequential execution costs without any parallelization or continuous batching:
\begin{equation}
\label{eq:latency_mai}
\text{Total Time}_{\text{MAI}} = \sum_{k \in \mathcal{C}} \left( \text{TTFT}_{k} + \frac{\#\text{Output Tokens}_{k}}{\text{Tok/s}_{k}} \right)
\end{equation}
where $\mathcal{C}$ is the set of sequential model calls triggered by the verification outcome in our Multi-Agent Inference (MAI) framework. $\text{TTFT}_{k}$ accounts for input preprocessing, vision encoding, and prefilling (dynamically looked up from \cref{tab:qwen_ttft} based on total context length), and $\text{Tok/s}_{k}$ is the average decoding throughput (\cref{tab:qwen_decoding}). 

By deliberately enforcing this fully sequential estimation, we isolate the fundamental algorithmic efficiency of our framework, ensuring that the speedups reported in the main text are robust and not artifacts of specific engineering optimizations.

\subsection{Routing Optimization via 2-Stage Decoding (S2).}
As motivated in \cref{sec:qwen_perf}, long reasoning chains can degrade the instruction-following capability of small models. When the 2B model generates the correct logical conclusion but outputs it in an incorrect format, it triggers an unnecessary mutual verification failure, forcing a fallback to the 8B model. Importantly, our core multi-agent framework effortlessly absorbs this: the 8B model, during Reasoning Transfer, receives this poorly formatted context, filters the noise, and outputs the perfectly formatted answer. 

However, to minimize these unnecessary 8B calls, we evaluate a cascade variant that replaces 2B-R with the two-stage decoding 2B-R(S2) (\cref{eq:2stage}) to serve as a \textit{routing optimization}. Empirically, this explicitly improves the verification success rate ($\mathrm{MV}_{2\mathrm{B\text{-}R}/8\mathrm{B\text{-}S}}$) and strategically reduces the reasoning-transfer (RT) rate across several benchmarks (\cref{tab:qwen3_routing}). Final cascade accuracy remains stable and even improves in some cases due to cleaner routing.

\begin{table*}[t]
    \centering
    \caption{%
        Original 2B-R vs.\ 2B-R(S2) in the cascade.%
        \emph{MV$|N$} matches the main table $\mathrm{MV}_{2\mathrm{B(R)}/8\mathrm{B(S)}}$ (count; \% of all tests).%
        \emph{RT$|N$} matches the main RT column (count; \% of all tests).%
        Final: POPE/F1, InfoVQA/mean ANLS$\times 100$, others/accuracy.%
    }
    \label{tab:qwen3_routing}
    \small
    \label{tab:qwen_2/8B_S2}
    \resizebox{\linewidth}{!}{%
    \begin{tabular}{l|r|ccc|ccc|r}
        \toprule
        \multirow{2}{*}{Dataset}
            & \multirow{2}{*}{\#Test}
            & \multicolumn{3}{c|}{2B-R}
            & \multicolumn{3}{c|}{2B-R(S2)}
            & \multirow{2}{*}{$\Delta$Final} \\
        \cmidrule(lr){3-5} \cmidrule(lr){6-8}
        & & MV$|N$ & RT$|N$ & Final & MV$|N$ & RT$|N$ & Final & \\
        \midrule
        POPE         & 9000  & 46 (0.51\%)  & 447 (4.97\%)  & 88.70 & 63 (0.70\%)  & 430 (4.78\%)  & 88.79 & +0.09 \\
        MMMU         & 900   & 141 (15.67\%) & 284 (31.56\%) & 61.22 & 63 (7.00\%)  & 362 (40.22\%) & 61.00 & $-0.22$ \\
        MMBench      & 4329  & 262 (6.05\%)  & 330 (7.62\%)  & 85.48 & 263 (6.08\%) & 329 (7.60\%)  & 85.48 & 0.00 \\
        ChartQA      & 2500  & 195 (7.80\%)  & 505 (20.20\%) & 86.80 & 216 (8.64\%) & 484 (19.36\%) & 86.80 & 0.00 \\
        InfoVQA      & 2801  & 271 (9.68\%)  & 802 (28.63\%) & 82.94 & 278 (9.93\%) & 795 (28.38\%) & 83.00 & +0.06 \\
        RealWorldQA  & 765   & 105 (13.73\%) & 151 (19.74\%) & 68.37 & 114 (14.90\%) & 142 (18.56\%) & 68.37 & 0.00 \\
        \bottomrule
    \end{tabular}%
    }
\end{table*}

\subsection{Latency Measurement and Experimental Setup}
\label{sec:append_latency_impl_details}

\paragraph{HuggingFace Experiments (SD and 8B-R Baseline).}
Both 8B-R and SD (8B+2B) use HuggingFace Transformers on a single H100. The 8B-R baseline runs standard autoregressive generation under the reason prompt. SD uses the identical prompt but replaces the decoding algorithm with assisted generation, where the 2B draft proposes tokens that the 8B target verifies. HuggingFace restricts assisted generation to batch size 1 (\texttt{ValueError} is raised for larger batches); the 8B-R baseline is evaluated under the same batch size 1 setting to ensure a strictly fair comparison. The reported latency covers data loading, preprocessing, and the full generation loop, excluding one-time model loading.

\paragraph{Step-level Profiling of Speculative Decoding.}
To quantitatively illustrate the context-switching bottlenecks discussed in Section 5.5, we profile a single ChartQA sample under the HuggingFace backend setup described above. We tested two different speculation lengths ($n$):
\begin{itemize}
    \item \textbf{$n=5$}: The 8B target is invoked 39 times. Out of 131 proposed tokens, 108 are accepted (17.56\% rejection rate), yielding a latency of 4.44s.
    \item \textbf{$n=20$}: The 8B target is invoked 30 times, but the longer draft sequences increase the rejection rate to 27.33\% (125/172 accepted), resulting in a latency of 4.19s.
\end{itemize}
In both configurations, the repeated invocations and high rejection rate negate the theoretical speedup of drafting, making the process strictly slower than the standard autoregressive baseline.

\paragraph{vLLM Experiments (MAI and 8B-R Baseline).}
Both MAI and the 8B-R baseline use vLLM with batch size 32 on a single H100. The 8B-R baseline runs on ChartQA(2,500 samples) with the reason prompt in a single vLLM process. MAI executes a three-stage cascade via subprocess orchestration: (1) 2B and 8B nothink subprocesses run in parallel on the same GPU, both producing short answers for all samples; (2) for the disagreement subset ($a_\text{2S} \neq a_\text{8S}$), a 2B subprocess runs reasoning; (3) for samples still in disagreement, an 8B subprocess reads the 2B reasoning context and outputs a short final answer. The reported latency is the end-to-end wall-clock time including all subprocess lifecycles (engine initialization, inference, and teardown). Note that vLLM does not currently support speculative decoding for VLMs, which is why SD is only evaluated under the HuggingFace backend.

\paragraph{MAI Implementation Status and Scalability.}
Our current implementation is a research prototype that restarts a fresh vLLM engine for every stage (Stage 1 nothink, Stage 2 reasoning, Stage 3 agent), incurring repeated cold-start costs including weight loading, KV-cache allocation, and CUDA graph capture. In a production setting, persistent vLLM engines with asynchronous stage handoff would eliminate these redundant startup overheads entirely. Despite this unoptimized implementation, MAI already achieves a 2.40$\times$ speedup over the 8B-R baseline (\cref{tab:sd_comparison}), indicating that the reported numbers represent a conservative lower bound.

Furthermore, our response-level design is naturally amenable to multi-GPU deployment: since models communicate only through complete text responses, the 2B and 8B models can be served on separate GPUs --- or even separate machines --- with negligible communication overhead. In contrast, token-level speculative decoding requires draft and target models to exchange predictions at every decoding step, tightly coupling them to shared memory or high-bandwidth interconnects.

\paragraph{Cross-group Note.}
Due to differences in backend and batching, latency is only comparable \textit{within} each group. Speedup is computed accordingly.

\subsection{Discussion and Practical Deployment} \label{sec:append_discussion}

\textbf{Memory.} 
For the 2B+8B configuration, the total parameter count is approximately 10B, fitting comfortably within a single H100 (96\,GB). The two models can also be served on separate GPUs or with time-sharing to reduce peak memory usage.

\textbf{Scalability.} 
Our response-level design is naturally amenable to multi-GPU deployment: since models communicate only through complete text responses, the 2B and 8B models can be served on separate GPUs or separate machines with negligible communication overhead. In contrast, token-level speculative decoding requires draft-target synchronization at every decoding step. Furthermore, our current prototype restarts a fresh vLLM engine per stage; persistent engines with asynchronous handoff would eliminate this overhead, making the reported $2.40\times$ speedup a conservative lower bound. See \cref{sec:append_latency_impl_details} for details.

\subsection{Qualitative Comparison} \label{sec:append_example}
To demonstrate the proposed multi-agent inference, we provide additional real examples to show the effectiveness of different components in our framework.

\paragraph{Qwen3-VL-2B-S vs. Qwen3-VL-2B-R}
First, we show the difference for the same Qwen3-VL-2B model with different prompts. From the example below, we can find that the 2B model cannot obtain the right answer with the simple prompt. By assigning more output token budgets for reasoning, the model can answer the question correctly, which confirms the benefit of additional output tokens for small models.

\begin{ExampleBox}[Example 1: More Output Tokens Helps]

The graph below shows the AD-AS diagram for Spain. All numbers are in billions. \\ [2pt]
\includegraphics[width=0.5\linewidth]{prompt_exp/eg1.png} \\[3pt]
What is the size of real GDP in the short-run equilibrium? \\[2pt]
A. 500 \\ 
B. 450 \\ 
C. 400 \\ 
D. 600 \\ [1pt]
Ground Truth: C

\begin{SimpleBox} 
\textbf{Prompt:} Without thinking, directly answer with the option's letter from the given choices directly in the format: [[answer]] \\ [2pt]
\textbf{Qwen3-VL-2B Answer:} [[A]] \WrongAns

\end{SimpleBox}

\begin{ReasonBox} 
\textbf{Prompt:} Please think step by step, provide a detailed explanation, then answer with the option's letter from the given choices directly in the format: [[answer]] \\

\textbf{Qwen3-VL-2B Answer:} \\[2pt]
To determine the size of real GDP in the short-run equilibrium, we need to identify the point where the Aggregate Demand (AD) curve intersects the Short-Run Aggregate Supply (SRAS) curve. \\[4pt]
Looking at the graph: \\[2pt]
- The \textbf{AD} curve is a downward-sloping line that starts at a price level of 1000 and a real GDP of 0. \\
- The \textbf{SRAS} curve is a rightward-sloping line that starts at a price level of 200 and a real GDP of 0. \\
- The \textbf{LRAS} (Long-Run Aggregate Supply) is a vertical line at 500 on the real GDP axis. \\[4pt]
In the short run, the equilibrium occurs at the intersection of AD and SRAS. We can see from the graph that the two curves intersect at a real GDP of 400. \\[4pt]
Therefore, the size of real GDP in the short-run equilibrium is 400. \\[4pt]
\text{[[C]]} \RightAns
\end{ReasonBox}

\end{ExampleBox}

\paragraph{Qwen3-VL-2B-R vs. Qwen3-VL-2B-R(S2)}
However, with more output tokens, it becomes challenging for small models to follow the instructions about the output format. Here is an example illustrating the issue. The right answer should be in ``Yes'' or ``No'' while the result after thinking returns in ``[[0]]''. With the proposed 2-stage strategy, \ie, (S2), we can obtain the answer in the right format with limited additional output tokens, \eg, in the following Example 2.

In particular, large-scale models maintain correct output formatting even after extended reasoning chains, effectively translating these longer thoughts into performance gains. In contrast, smaller models often encounter formatting errors. Consequently, this two-stage strategy is introduced to mitigate the formatting deficiencies observed in smaller models during the assessment of their reasoning capabilities, thereby ensuring that their true reasoning performance can be measured and compared objectively

\begin{ExampleBox}[Example 2: 2-Stage Mitigates Format Issue]

The image shows a python code. Is the output of the code `1'? \\
\includegraphics[width=0.4\linewidth]{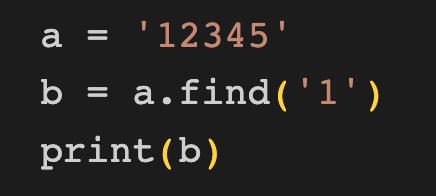} \\
Ground Truth: No

\begin{ReasonBox} 
\textbf{Prompt:} Please think step by step, provide a detailed explanation, then answer the question using a single word or phrase in the format: [[answer]] \\

\textbf{Qwen3-VL-2B Answer:} \\[2pt]
Let's analyze the code step by step:\\ \\
1. The variable `a` is assigned the string value `'12345'`.\\ 
2. The line `b = a.find('1')` calls the `find` method on the string `a`. The `find` method returns the index of the first occurrence of the substring '1' in the string `a`, or -1 if the substring is not found.\\ 
3. In the string `'12345'`, the character '1' appears at the very beginning, so its index is 0.\\ 
4. Therefore, `b` will be assigned the value `0`.\\ 5. The final line `print(b)` will print the value of `b`, which is `0`.\\ \\
The output of the code is not '1', but rather the integer `0`.\\ [2pt]
\text{[[0]]} {{\color{red!70!black} \faTimesCircle\ Wrong\ answer format}}

\end{ReasonBox}

\begin{Stage2Box}[Stage2: R(S2)] 
\textbf{Prompt:} According to previous thinking:\{Stage1 Response\}. \\
Answer the question using a single word or phrase.\\
 
\textbf{Qwen3-VL-2B Answer:} No {{\color{green!60!black} \faCheckCircle\ Right\ answer format}}
\end{Stage2Box}

\end{ExampleBox}

\paragraph{Reasoning Transfer between 2B and 8B Models}
There are some challenging examples that both 8B-S and 2B-R cannot handle well, which may require the reasoning from the large model. Here is an example as follows. We find that 8B-S and 2B-R show the wrong answer, while 8B-R returns the appropriate result. However, it costs a lot of output tokens from the 8B model, which can be slow. Therefore, we try to transfer the reasoning tokens to the 8B model as input tokens, which will not significantly increase the latency according to our analysis with simulation. With the simple prompt, the 8B model can return the right answer with transferred reasoning tokens, which confirms our proposal.

\begin{ExampleBox}[\fontsize{9.6pt}{13pt}\selectfont  Example 3: \text{Challenging Case for Reasoning Transfer}]

I feel very cold. Is the tool in the picture suitable for use?  \\
\includegraphics[width=0.2\linewidth]{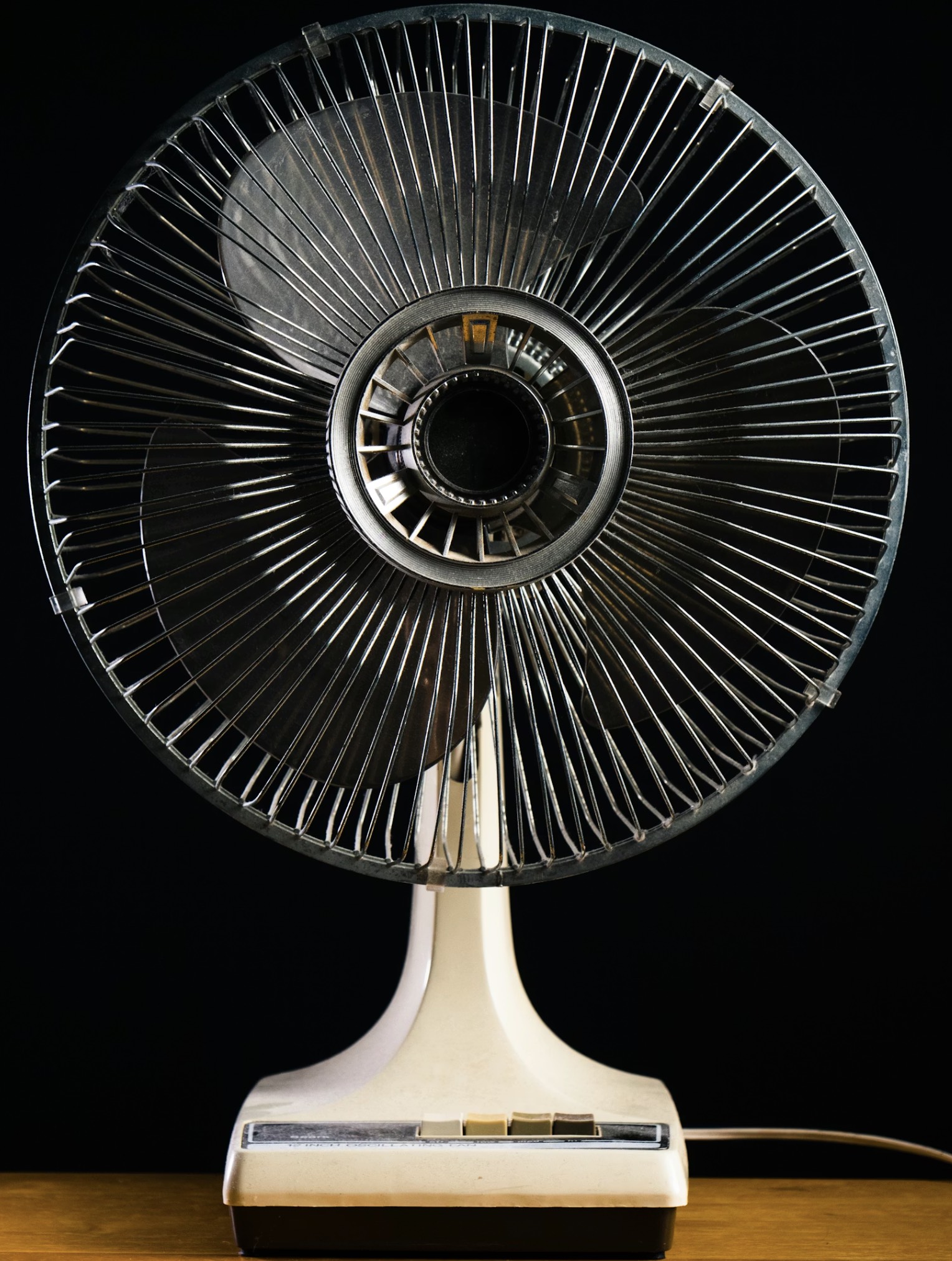} 

Ground Truth: No

\begin{SimpleBox}[Simple (Qwen3-VL-8B)] 
\textbf{Prompt:} Without thinking, directly answer the question using a single word or phrase in the format: [[answer]]

\textbf{Qwen3-VL-8B Answer:} [[Yes]] \WrongAns
\end{SimpleBox}

\begin{ReasonBox} [Reasoning (Qwen3-VL-2B) ] 
\textbf{Prompt:} Please think step by step, provide a detailed explanation, then answer the question using a single word or phrase in the format: [[answer]] \\

\textbf{Qwen3-VL-2B Answer:} \\[2pt]
Let's analyze the image and the question step by step.\\1.  \textbf{Identify the Object}: The image shows a fan, specifically a table fan, which is a device designed to {\color{red}circulate air and cool a room or a person}.\\2.  \textbf{Analyze the Function}: The primary function of a fan is to move air. This is done by creating a flow of air, which can help to {\color{red}lower the temperature of a room or a person}.\\3.  \textbf{Assess the Current Situation}: The user states they are "very cold." This suggests they are {\color{red}experiencing cold conditions}, possibly in a room with low temperatures or in a cold environment.\\4.  \textbf{Evaluate the Suitability}: Since {\color{red}the fan is designed to cool down a room or a person}, and the user is feeling cold, {\color{red}the fan would be suitable for use to help warm up the environment and the user}. It would be a practical tool to help combat the cold.\\5.  \textbf{Conclusion}: The fan is a tool designed for cooling, which is the opposite of what the user is feeling. Therefore, it is suitable for use to help them feel warmer.\\
[4pt]
[[yes]] {{\color{red!70!black} \faTimesCircle\ Wrong\ answer. Reasoning error.}} \\

------------------------------------------------------------ \\
\textbf{ \faClock\ Running Time:} 3.83 s 
\end{ReasonBox}

\begin{ReasonBox}[Reasoning (Qwen3-VL-8B) ] 
\textbf{Prompt:} Please think step by step, provide a detailed explanation, then answer the question using a single word or phrase in the format: [[answer]] \\

\textbf{Qwen3-VL-8B Answer:} \\ [2pt]
Let's think step by step:\\ 
1.  The object in the picture is a vintage-style oscillating table fan. Its primary function is to circulate air and create a cooling breeze, which helps to lower the perceived temperature in a room.\\ 
2.  The user states they feel "very cold." This indicates a {\color{green!60!black} need for warmth} or at least a reduction in the sensation of cold.\\ 
3.  A fan works by moving air, which can make a person feel cooler, especially if the ambient temperature is already low or if the air is moving across the skin. {\color{green!60!black} It does not generate heat.}\\
4.  Therefore, using a fan when one feels very cold would likely make the situation worse, as it would increase the air movement and potentially {\color{green!60!black} make the person feel even colder.}\\
5.  The tool is not designed to provide warmth. It is a cooling device.\\
Conclusion: The tool shown in the picture is {\color{green!60!black} not suitable for use when one feels very cold}, as it will exacerbate the feeling of coldness rather than alleviate it.\\ [2pt]
[[No]] {{\color{green!60!black} \faCheckCircle\ Right\ answer}}

------------------------------------------------------------ \\

\textbf{ \faClock\ Running Time:} 4.56 s

\end{ReasonBox}
 
\begin{TransferBox}[Reason Transfer: 8B + 2B-R] 
\textbf{Prompt:} The following is a reference answer from a smaller model (it might be
wrong or incomplete): {\color{reason_color} Let's analyze the image and the question step by step.\textbackslash n\textbackslash n1.  **Identify the Object**: The image shows a fan, specifically a table fan, which is a device designed to circulate air and cool a room or a person.\textbackslash n\textbackslash n2.  **Analyze the Function**: The primary function of a fan is to move air. This is done by creating a flow of air, which can help to lower the temperature of a room or a person.\textbackslash n\textbackslash n3.  **Assess the Current Situation**: The user states they are "very cold." This suggests they are experiencing cold conditions, possibly in a room with low temperatures or in a cold environment.\textbackslash n\textbackslash n4.  **Evaluate the Suitability**: Since the fan is designed to cool down a room or a person, and the user is feeling cold, the fan would be suitable for use to help warm up the environment and the user. It would be a practical tool to help combat the cold.\textbackslash n\textbackslash n5.  **Conclusion**: The fan is a tool designed for cooling, which is the opposite of what the user is feeling. Therefore, it is suitable for use to help them feel warmer.} \textbackslash n\textbackslash n   \textbackslash n\textbackslash n
Please answer the question yourself, using the reference only if it seems reasonable. \textbackslash n
Answer the question using a single word or phrase. \\[4pt]
\textbf{Qwen3-VL-8B Answer:} No {{\color{green!60!black} \faCheckCircle\ Right\ answer}} \\

------------------------------------------------------------ \\

\textbf{ \faClock\ Running Time:} 3.83 s + 0.126 s = 3.96 s 

\end{TransferBox}

\end{ExampleBox}

We also report the running time estimated as follows:
\begin{equation}
\text{Total Time (s)}_{\text{S,E,R}} = \text{TTFT (s)} + \frac{\text{\#Output Tok}}{\text{\#Tok/s}} 
\end{equation}

\begin{equation}
\begin{aligned}
\text{Total Time (s)}_{\text{8B+2B(R)}}
&= \text{TTFT (s)}_{\text{2B}} + \frac{\text{\#Output Tok}_{\text{2B}}}{\text{\#Tok/s}_{\text{2B}}} \\
& + \text{TTFT (s)}_{\text{8B}} + \frac{\text{\#Output Tok}_{\text{8B}}}{\text{\#Tok/s}_{\text{8B}}} 
\end{aligned}
\end{equation}
Note: $TTFT(s)_{8B}$ is determined via dynamic look-up based on the total context length (profiled up to 8192 tokens), ensuring accurate estimation for variable-length reasoning transfer.

From this example, we can find that 8B-R uses 5.54s to obtain the right answer while 2B-R costs 2.02s to generate reasoning tokens. By transferring the tokens to 8B-S, the combination can handle the example well with only about 2.03s, which is more efficient than 8B-R.

\subsection{Sparsity in Reasoning Token Transfer} \label{sec:append_sparsity}
\label{sec:sparse}
As shown in Corollary~1, the success of reasoning transfer is from the sparsity in self-attention operations. To validate our theoretical analysis, we study the attention scores for reasoning tokens from different models with the example above.

\begin{figure*}[t]
    \centering
        \begin{subfigure}{0.48\linewidth}
        \centering
        \includegraphics[width=\linewidth]{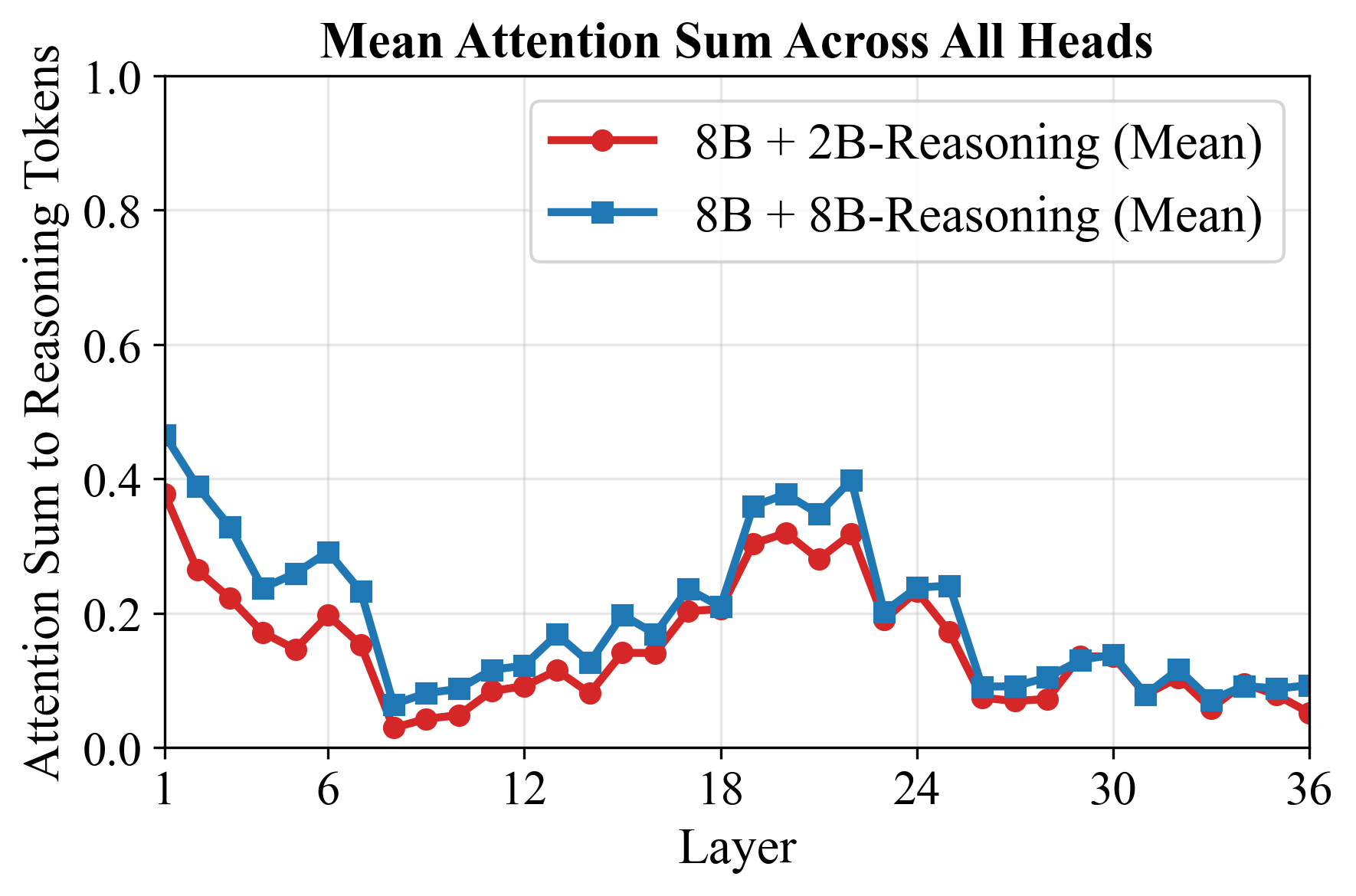}
        \caption{Attention weights of all reasoning tokens}
        \label{fig:reasoning_attention_layers:a}
    \end{subfigure}
    \centering
    \begin{subfigure}{0.48\linewidth}
    \centering
    \includegraphics[width=\linewidth]{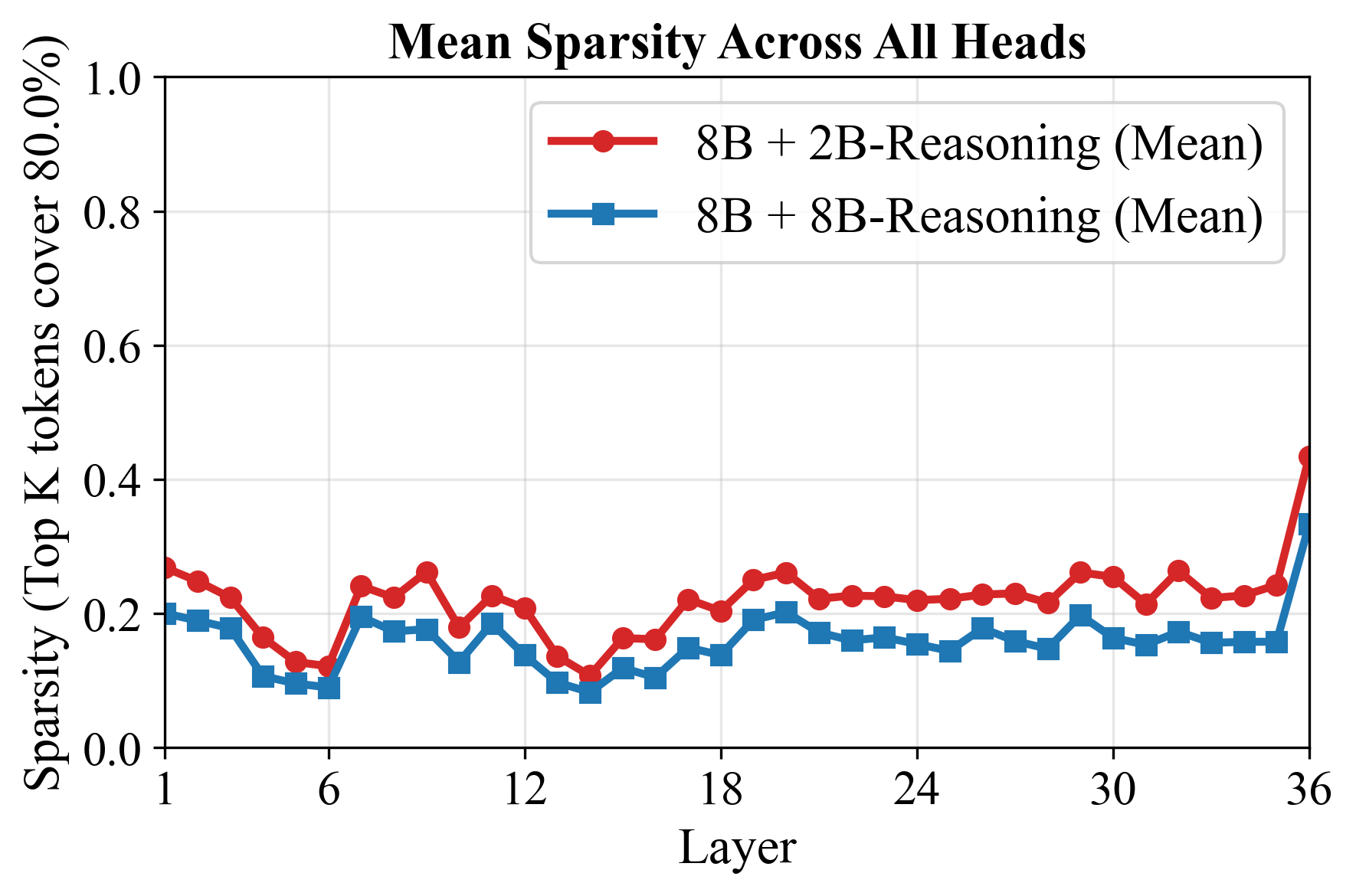}
    \caption{Sparsity within reasoning tokens}
    \label{fig:reasoning_attention_layers:b}
\end{subfigure}

    \caption{ Illustration of total attention weights of all reasoning tokens and sparsity within reasoning tokens averaged over 32 heads. Sparsity is measured by the ratio of reasoning tokens that contribute to 80\% total attention weights of reasoning tokens.
    }
    \label{fig:reasoning_attention_layers}
\end{figure*}

\begin{figure*}
    \centering
    \includegraphics[width=0.98\linewidth]{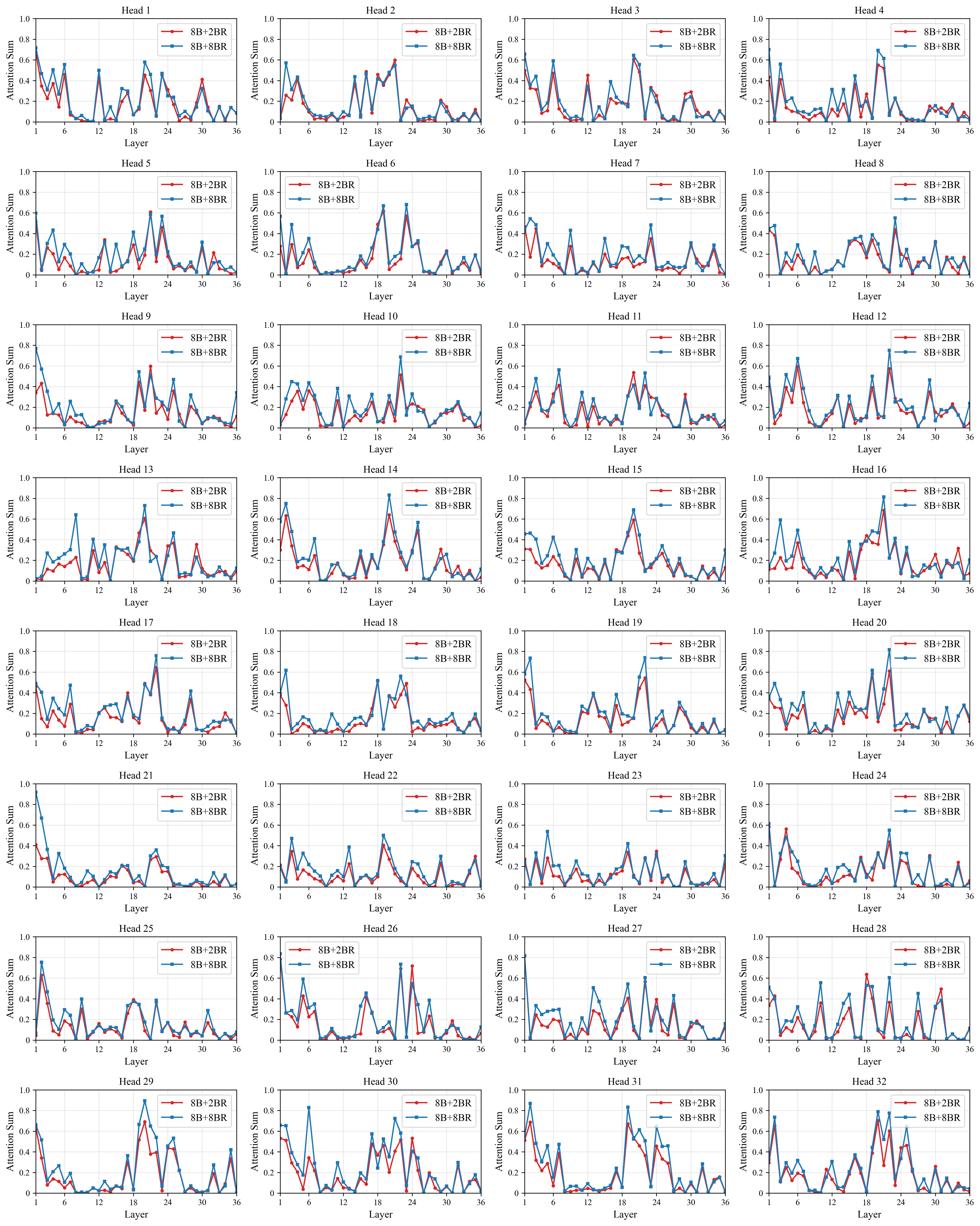}
 
    \caption{Head-wise attention weights of total reasoning tokens across different layers. 
     }
    \label{fig:reasoning_attention_layers_sum}
\end{figure*}

\begin{figure*}
    \centering
    \includegraphics[width=0.98\linewidth]{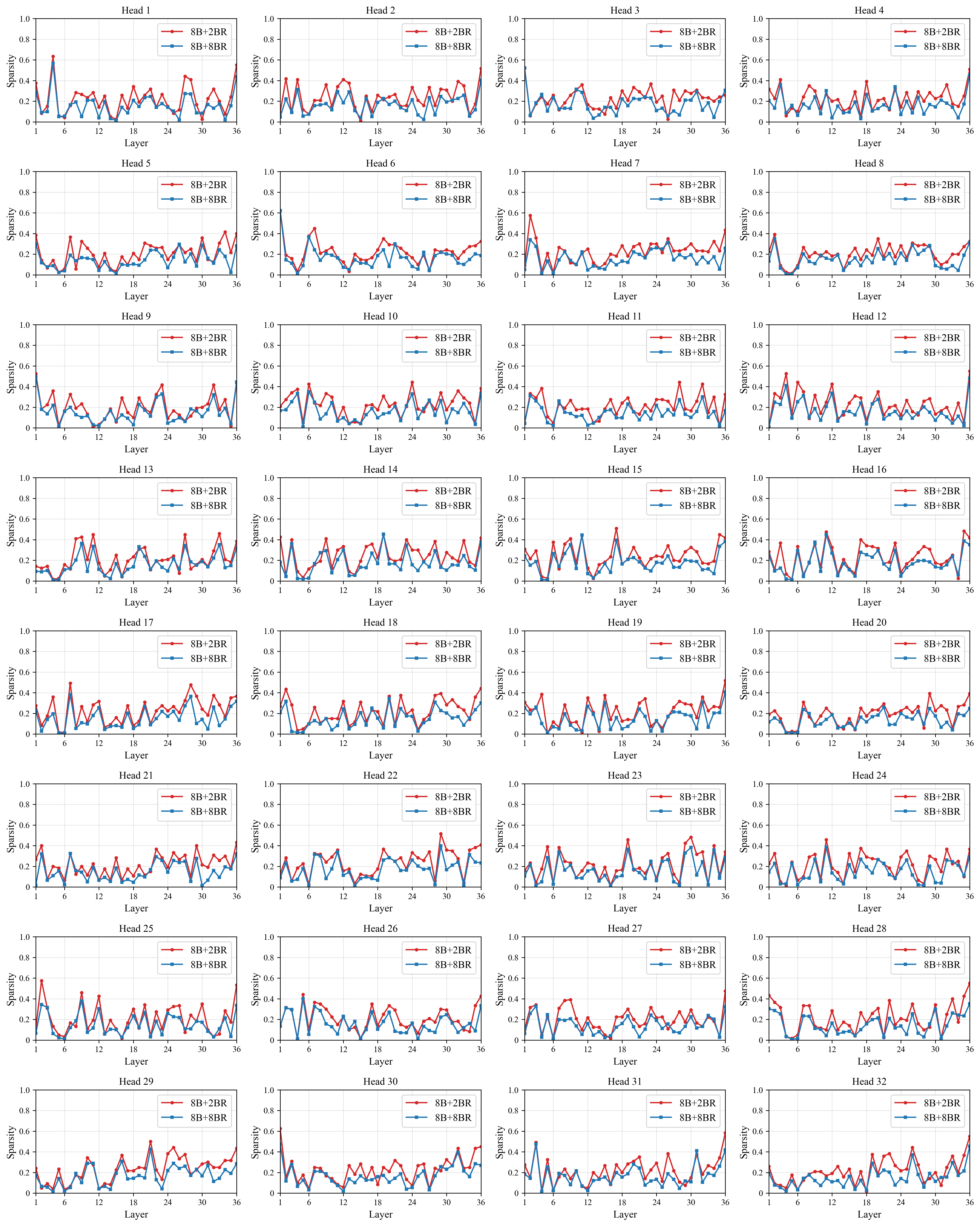}
 
    \caption{Head-wise sparsity within reasoning tokens across different layers.}  
    \label{fig:reasoning_attention_layers_sparse}
\end{figure*}

First, we show the distribution of attention scores over different layers in Qwen3-VL-8B model in \cref{fig:reasoning_attention_layers}, where \cref{fig:reasoning_attention_layers:a} shows the total weight of all reasoning tokens and \cref{fig:reasoning_attention_layers:b} shows the sparsity within reasoning tokens. The reported results are averaged over 32 heads and the detailed distribution for each head can be found in \cref{fig:reasoning_attention_layers_sum} and \cref{fig:reasoning_attention_layers_sparse}. We can find that with reasoning tokens from 8B-R and 2B-R respectively, the distribution is similar. For most of layers, the distribution within reasoning tokens is very skewed, where about the top 20\% reasoning tokens contribute to 80\% of the weights, and it confirms our analysis. For the last layer, the attention weights are mainly distributed on non-reasoning tokens for generation, \eg, answer prompt, and the internal distribution of reasoning tokens can be more balanced due to the smoothness of softmax operators. 

To demonstrate the selected tokens, we list the top-ranked reasoning tokens for different layers from 8B-R and 2B-R in \cref{tab:attn_comparison_merged}. The weight of each token is averaged over all heads before ranking. Obviously, the deeper layers focus more on semantic information from reasoning tokens. For these layers, we observe that the similar tokens, \eg, ``picture'', ``tool'', are picked by the 8B model from different reasoning tokens by Layer 24. It shows that key tokens are more important in inference than other reasoning tokens. It implies that even the sparse reasoning tokens from small models can help large models to think.

\begin{table*}[h!]
    \centering
\caption{Comparison of top tokens selected by \textbf{8B + 2B-R} and \textbf{8B + 8B-R} across different layers. Evidently, deeper layers focus on semantic reasoning tokens, where similar tokens can be selected from 2B-R and 8B-R tokens.}
    \label{tab:attn_comparison_merged}
\resizebox{\linewidth}{!}{
\begin{tabular}{c|c|c|c|c|c|c|c|c|c|c|c|c|c|c|c|c|c}
\toprule
\textbf{Layer} & \textbf{Model} & \textbf{N-Tok (80\%)} & \textbf{1st} & \textbf{2nd} & \textbf{3rd} & \textbf{4th} & \textbf{5th} & \textbf{6th} & \textbf{7th} & \textbf{8th} & \textbf{9th} & \textbf{10th} & \textbf{11th} & \textbf{12th} & \textbf{13th} & \textbf{14th} & \textbf{15th} \\
\midrule
\multirow{4}{*}{L1} & \multirow{2}{*}{8B+2BR} & \multirow{2}{*}{45.0\% (54)} & . & a & , & . & , & , & , & the & in & the & , & the & and & . & ) \\
& & & 11.1\% & 3.1\% & 3.0\% & 2.9\% & 2.9\% & 2.9\% & 2.6\% & 2.6\% & 2.4\% & 2.3\% & 2.1\% & 2.0\% & 2.0\% & 1.8\% & 1.7\% \\
\cmidrule(lr){2-18}
& \multirow{2}{*}{8B+8BR} & \multirow{2}{*}{40.6\% (104)} & . & . & the & , & of & . & ). & : & , & , & the & . & , & in & and \\
& & & 7.5\% & 3.5\% & 3.2\% & 2.8\% & 2.2\% & 2.2\% & 2.1\% & 2.0\% & 2.0\% & 1.8\% & 1.7\% & 1.5\% & 1.4\% & 1.4\% & 1.3\% \\
\midrule
 
\multirow{4}{*}{L6} & \multirow{2}{*}{8B+2BR} & \multirow{2}{*}{21.7\% (26)} & the & . & the & the & . & . & . & \textcolor{red}{Therefore} & . & , & the & The & \textcolor{red}{picture} & , & the \\
& & & 19.1\% & 17.6\% & 16.2\% & 3.7\% & 3.5\% & 3.3\% & 1.6\% & 1.5\% & 1.2\% & 1.0\% & 0.9\% & 0.9\% & 0.9\% & 0.9\% & 0.8\% \\
\cmidrule(lr){2-18}
& \multirow{2}{*}{8B+8BR} & 16.4\% (42) & the & the & . & the & : & . & . & ). & : & . & \textcolor{red}{Conclusion} & . & 2 & 5 & The \\
& & & 21.3\% & 14.8\% & 6.0\% & 4.5\% & 4.0\% & 2.8\% & 2.5\% & 2.2\% & 2.0\% & 1.9\% & 1.8\% & 1.7\% & 1.1\% & 0.8\% & 0.7\% \\
 \midrule
\multirow{4}{*}{L12} & \multirow{2}{*}{8B+2BR} & \multirow{2}{*}{29.2\% (35)} & . & \textcolor{red}{tool} & . & \textcolor{red}{tool} & the & The & . & \textcolor{red}{Therefore} & The & , & . & \textcolor{red}{provided} & \textcolor{red}{picture} & \textcolor{red}{picture} & , \\
& & & 18.2\% & 8.7\% & 6.7\% & 5.2\% & 4.3\% & 3.4\% & 3.2\% & 2.6\% & 2.3\% & 2.1\% & 2.0\% & 1.5\% & 1.4\% & 1.2\% & 1.2\% \\
\cmidrule(lr){2-18}
& \multirow{2}{*}{8B+8BR} &  \multirow{2}{*}{17.6\% (45)} & : & . & \textcolor{red}{step} & \textcolor{red}{think} &  & . & 2 & . & . & . & . & \textcolor{red}{Conclusion} & : &  & ). \\
& & & 14.1\% & 8.2\% & 7.8\% & 3.2\% & 2.9\% & 2.7\% & 2.5\% & 2.4\% & 2.3\% & 2.2\% & 2.1\% & 2.1\% & 2.0\% & 1.9\% & 1.6\% \\
\specialrule{0.5pt}{3pt}{1pt} \specialrule{0.5pt}{1pt}{3pt}
\multirow{4}{*}{L18} & \multirow{2}{*}{8B+2BR} & \multirow{2}{*}{28.3\% (34)} & . & the & \textcolor{red}{suitable} & , & , & \textcolor{red}{use} & \textcolor{red}{suitable} & , & which & . & The & the & \textcolor{red}{tool} & The & \textcolor{red}{However} \\
& & & 12.4\% & 4.8\% & 4.8\% & 3.7\% & 3.3\% & 3.2\% & 3.0\% & 2.9\% & 2.6\% & 2.6\% & 2.4\% & 2.2\% & 2.2\% & 2.1\% & 2.0\% \\
\cmidrule(lr){2-18}
& \multirow{2}{*}{8B+8BR} &  \multirow{2}{*}{21.1\% (54)} & . & , & : &  \textcolor{red}{uns} & . & such &  & . & . & ." & : & The & 2 & . & \textcolor{red}{very} \\
& & & 7.1\% & 5.4\% & 4.2\% & 3.0\% & 3.0\% & 2.7\% & 2.6\% & 2.6\% & 2.4\% & 2.2\% & 1.9\% & 1.9\% & 1.9\% & 1.8\% & 1.8\% \\
\midrule
 
\multirow{4}{*}{L24} & \multirow{2}{*}{8B+2BR} &  \multirow{2}{*}{24.2\% (29)} & . & \textcolor{red}{suitable} & \textcolor{red}{picture} & \textcolor{red}{use} & , & is & the & \textcolor{red}{tool} & for & in & \textcolor{red}{suitable} & , & . & the & \textcolor{red}{Therefore} \\
& & & 12.4\% & 9.9\% & 4.6\% & 4.2\% & 4.1\% & 3.2\% & 3.1\% & 3.0\% & 2.7\% & 2.6\% & 2.6\% & 2.4\% & 2.2\% & 2.2\% & 2.1\% \\
\cmidrule(lr){2-18}
& \multirow{2}{*}{8B+8BR} &  \multirow{2}{*}{21.1\% (54)} & \textcolor{red}{suitable} & . & \textcolor{red}{uns} & is & The & \textcolor{red}{not} &  \textcolor{red}{use} & : & \textcolor{red}{uitable} & , & \textcolor{red}{tool} & ) & . & the & \textcolor{red}{picture} \\
& & & 7.9\% & 5.9\% & 4.5\% & 3.9\% & 3.7\% & 3.3\% & 2.9\% & 2.9\% & 2.8\% & 2.4\% & 2.3\% & 2.1\% & 2.1\% & 1.9\% & 1.6\% \\
\midrule
 
\multirow{4}{*}{L30} & \multirow{2}{*}{8B+2BR} &  \multirow{2}{*}{27.5\% (33) }& \textcolor{red}{suitable} & \textcolor{red}{fan} & . & is & The & a & \textcolor{red}{suitable} & , & the & , & the & \textcolor{red}{picture} & \textcolor{red}{tool} & \textcolor{red}{cold} & \textcolor{red}{picture} \\
& & & 11.0\% & 9.2\% & 7.0\% & 4.4\% & 4.1\% & 3.9\% & 3.8\% & 3.2\% & 2.7\% & 2.7\% & 1.9\% & 1.8\% & 1.7\% & 1.7\% & 1.6\% \\
\cmidrule(lr){2-18}
& \multirow{2}{*}{8B+8BR} &  \multirow{2}{*}{25.0\% (64)}  & \textcolor{red}{not} & \textcolor{red}{suitable} & \textcolor{red}{uitable} & is & The &  \textcolor{red}{uns} & . &  & \textcolor{red}{step} & : & \textcolor{red}{cold} & . & \textcolor{red}{Conclusion} & , & . \\
& & & 9.2\% & 5.4\% & 4.3\% & 4.1\% & 4.0\% & 3.6\% & 3.4\% & 3.2\% & 2.1\% & 2.1\% & 1.9\% & 1.8\% & 1.6\% & 1.6\% & 1.4\% \\
\specialrule{0.5pt}{3pt}{1pt} \specialrule{0.5pt}{1pt}{3pt}
\multirow{4}{*}{L36} & \multirow{2}{*}{8B+2BR} &  \multirow{2}{*}{53.3\% (64)} & . & , & , & \textcolor{red}{fan} & the & . & . & the & is & in & . & a & \textcolor{red}{Therefore} & a & the \\
& & & 7.5\% & 5.3\% & 3.3\% & 3.0\% & 2.7\% & 2.6\% & 2.1\% & 2.1\% & 2.0\% & 1.8\% & 1.7\% & 1.7\% & 1.7\% & 1.6\% & 1.6\% \\
\cmidrule(lr){2-18}
& \multirow{2}{*}{8B+8BR} &  \multirow{2}{*}{46.9\% (120)} &  & . & . & : & 1 & ). & \textcolor{red}{Conclusion} & . & is & \textcolor{red}{cold} & : & such &  & \textcolor{red}{step} & \textcolor{red}{fan} \\
& & & 4.8\% & 3.8\% & 2.6\% & 2.4\% & 1.7\% & 1.7\% & 1.7\% & 1.6\% & 1.6\% & 1.5\% & 1.4\% & 1.4\% & 1.4\% & 1.3\% & 1.0\% \\
\bottomrule
\end{tabular}
}
\end{table*}

\section{All Prompt Templates}

We list all prompts used in our experiments as follows. Following the official setting of the LMMS-Eval Framework, we set “answer the question using a single word or phrase” for POPE, and “answer with the option’s letter from the given choices directly” for the other benchmarks.

\begin{SimpleBox}[Simple (S)]
\textbf{Prompt1:} \textbf{Without thinking, directly} answer the question using a single word or phrase in the format: [[answer]] \\

\textbf{Prompt2:} \textbf{Without thinking, directly} answer with the option's letter from the given choices in the format: [[answer]]. 
\end{SimpleBox}

\begin{ExplainBox}[Explain (E)]
\textbf{Prompt 1:} \textbf{Provide a simple explanation, then} answer the question using a single word or phrase in the format: [[answer]] \\

\textbf{Prompt 2:} \textbf{Provide a simple explanation, then} answer with the option's letter from the given choices directly in the format: [[answer]] 
\end{ExplainBox}

\begin{ReasonBox}[Reasoning (R)]
\textbf{Prompt 1:} \textbf{Please think step by step, provide a detailed explanation, then} answer the question using a single word or phrase in the format: [[answer]] \\

\textbf{Prompt 2:} \textbf{Please think step by step, provide a detailed explanation, then} answer with the option's letter from the given choices directly in the format: [[answer]] 
\end{ReasonBox}

\begin{Stage2Box}[Stage 2]
\textbf{Prompt 1:} According to previous thinking: \{stage1\_response\}\textbackslash n
Answer the question using a single word or phrase. \\

\textbf{Prompt 2:} According to previous thinking: \{stage1\_response\}\textbackslash n
Answer with the option's letter from the given choices directly. 
\end{Stage2Box}

\begin{TransferBox}[Reason Transfer]
For POPE and ChartQA, \\ [1pt]
\textbf{Prompt 1:} According to previous thinking: \{stage1\_response\}\textbackslash n
Answer the question using a single word or phrase.  \\

For MMBench, MMMU and InfoVQA, RealWorldQA, \\ [1pt]
\textbf{Prompt 2:} The following is a reference answer from a smaller model
(it might be wrong or incomplete): \{stage1\_response\}\textbackslash n\textbackslash n\textbackslash n\textbackslash n
Please answer the question yourself, using the reference only if it seems reasonable.\textbackslash n
Answer with the option's letter from the given choices directly in the format: [[answer]].
\end{TransferBox}



\end{document}